\documentclass{article}

\PassOptionsToPackage{numbers, compress}{natbib}
\usepackage[final]{neurips_2025}

\usepackage{subfloat}
\usepackage{enumitem} % 引入enumitem宏包
\usepackage{amsmath}
\usepackage{graphicx} 
\usepackage[utf8]{inputenc} % allow utf-8 input
\usepackage[T1]{fontenc}    % use 8-bit T1 fonts
\usepackage{hyperref}       % hyperlinks
\usepackage{url}            % simple URL typesetting
\usepackage{booktabs}       % professional-quality tables
\usepackage{amsfonts}       % blackboard math symbols
\usepackage{nicefrac}       % compact symbols for 1/2, etc.
\usepackage{microtype}      % microtypography
\usepackage{xcolor}         % colors
\usepackage{subcaption}
\usepackage{caption}
\usepackage{multirow}
\usepackage{wrapfig} % 引入wrapfig宏包
\usepackage{fontawesome}  % 提供专业矢量图标
\newcommand{\cmark}{\textcolor{black}{\faCheck}}   % ✓
\newcommand{\xmark}{\textcolor{black}{\faTimes}}    % ✗

\title{PipeFusion: Patch-level Pipeline Parallelism for Diffusion Transformers Inference}

\author{%
  Jiarui Fang \thanks{Equal contribution.}  \thanks{This work was completed when Jiarui Fang, Jinzhe Pan, Aoyu Li, and Jiannan Wang were with Tencent.} \\
  ByteDance \\
  \texttt{fangjiarui123@gmail.com} \\
  \And
  Jinzhe Pan\footnotemark[1] \footnotemark[2]\\
  HUST \\
  \texttt{eigensystem@hust.edu.cn} \\
  \And
  Aoyu Li\footnotemark[2] \\
  ByteDance \\
  \texttt{aoyuli2000@outlook.com} \\
  \And
  Xibo Sun \\
  Tencent \\
  \texttt{xibosun@tencent.com} \\
  \And
  Jiannan Wang\footnotemark[2] \\
  The University of Hong Kong \\
  \texttt{steaunk@connect.hku.hk} \\
}

\begin{document}

\maketitle

\begin{abstract}
This paper presents PipeFusion, an innovative parallel methodology to tackle the high latency issues associated with generating high-resolution images using diffusion transformers (DiTs) models. 
PipeFusion partitions images into patches and the model layers across multiple GPUs. It employs a patch-level pipeline parallel strategy to orchestrate communication and computation efficiently. 
By capitalizing on the high similarity between inputs from successive diffusion steps, PipeFusion reuses one-step stale feature maps to provide context for the current pipeline step.
This approach notably reduces communication costs compared to existing DiTs inference parallelism, including tensor parallel, sequence parallel and DistriFusion.
PipeFusion enhances memory efficiency through parameter distribution across devices, ideal for large DiTs like Flux.1.
Experimental results demonstrate that PipeFusion achieves state-of-the-art performance on 8$\times$L40 PCIe GPUs for Pixart, Stable-Diffusion 3, and Flux.1 models.
Our source code is available at \url{https://github.com/xdit-project/xDiT}.
\end{abstract}

\section{Introduction}
In recent years, the capabilities of AI-generated content (AIGC) have rapidly advanced, with diffusion models emerging as a leading generative technique in image~\cite{blackforestlabs2023announcement, esser2024scaling, chen2024pixart, peebles2023scalable} and video synthesis~\cite{sora2024, meta2024moviegen}. 
The model architecture of diffusion models is undergoing a significant transformation.
Traditionally dominated by U-Net~\cite{ronneberger2015u} architectures, these models are now transitioning to Diffusion Transformers (DiTs)~\cite{peebles2023scalable}. 
The context length of diffusion models is also increasing, which is crucial for handling long videos and high-resolution images.

DiTs suffer from quadratic latency growth in long-sequence generation due to attention computation.
Given that a single GPU cannot satisfy the latency requirements for practical applications, it becomes necessary to parallelize the DiTs inference for a single image across multiple computational devices.
However, tensor parallelism (TP)~\cite{shoeybi2019megatron}, commonly used in serving Large Language Models (LLMs), is inefficient for DiTs due to their large activation volumes. 
In such cases, the communication costs often outweigh the benefits of parallel computation.
To address the challenges posed by long sequences, sequence parallelism (SP)~\cite{jacobs2023deepspeed, liu2023ring} is applied for the parallel inference of DiTs, which partitions the input image into patches and needs data exchange across different devices in the attention operations.

Recent studies have highlighted the presence of \textit{input temporal redundancy} in Diffusion Transformers (DiTs) during inference~\cite{so2024frdiff, ma2024deepcache, ma2024learning, yuan2024ditfastattn, zhao2024real, meta2024moviegen}. This phenomenon manifests as significant similarity in both inputs and activations across consecutive diffusion time steps, presenting opportunities for optimizing parallel computation approaches. DistriFusion~\cite{li2023distrifusion} represents a sequence parallelism method that exploits this characteristic in U-Net-based diffusion models. The approach combines fresh local activations with stale activations from preceding steps during attention and convolution operations, effectively overlapping communication costs within individual diffusion timestep computations. However, applying DistriFusion to DiTs reveals substantial memory inefficiencies, as the method requires maintaining full spatial dimensions of attention keys (K) and values (V) across all layers, resulting in memory costs that scale linearly with sequence length.

To more effectively utilize input temporal redundancy, we propose \textbf{PipeFusion}, a novel patch-level pipeline parallelism.
PipeFusion partitions both the DiTs layers and input image patches across GPUs, enabling pipelined parallel processing by leveraging temporal redundancy to reuse stale activations from previous timesteps.
Firstly,  PipeFusion substantially reduces inter-device communication costs compared to existing DiTs' parallel methods. It only transfers the input of the initial layer and the output of the final layer on each device, whereas other parallel approaches, i.e. sequence and tensor parallel, require transmitting activations for every DiT layer. 
Secondly, PipeFusion offers lower memory consumption by distributing model parameters across multiple devices.
This makes it particularly suitable for DiTs with large model parameters, such as the 12B Flux.1 model~\cite{blackforestlabs2023announcement}. 
In contrast to DistriFusion, PipeFusion reduces the memory buffers required for attention K, V to 1/$N$ ($N$ is the parallel degree). 
Finally, PipeFusion, by utilizing fresh K and V values over a longer period, achieves higher image generation accuracy compared to DistriFusion.
Experimental results show that PipeFusion achieves the lowest DiTs inference latency on 8$\times$L40 PCIe-connected GPUs compared to sequence parallelism, DistriFusion, and tensor parallelism.

\section{Background \& Related Works}

\begin{figure}[htbp!]
\centering
\includegraphics[width=0.9\textwidth]{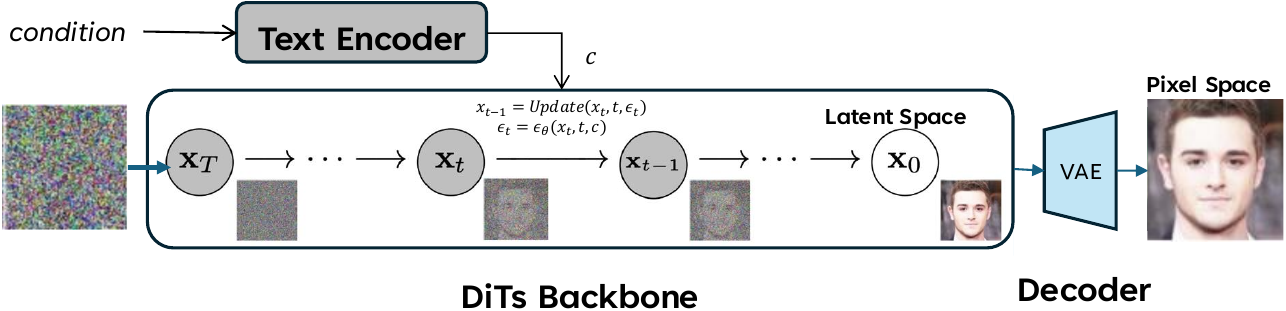}
\caption{Workflow of DiTs inference.}
\label{fig:ditinfer}
\end{figure}

\textbf{Diffusion Models and Diffusion Transformers (DiTs)}: Diffusion models generate high-quality images using a noise-prediction deep neural network (DNN) denoted by $\epsilon_{\theta}$. The process begins with pure Gaussian noise $x_T \sim \mathcal{N}(0, I)$ and iteratively denoises it through multiple steps to produce the final image $x_0$, with $T$ being the total number of diffusion time steps. 
As shown in Figure~\ref{fig:ditinfer}, at each step $t$, given the noisy image $x_t$, the model $\epsilon_{\theta}$ takes $x_t$, $t$, and an additional condition $c$ (e.g., text, image) as inputs to predict the noise $\epsilon_t$ within $x_t$. The previous image $x_{t-1}$ is obtained using:
\begin{equation}
    x_{t-1} = \text{Update}(x_t, t, \epsilon_t), \quad \epsilon_t = \epsilon_{\theta}(x_t, t, c).
\end{equation}
The function \text{Update} is specific to the sampler (e.g., DDIM~\cite{song2020denoising}, DPM~\cite{lu2022dpm}) and involves element-wise operations. Finally, $x_0$ is decoded from the Latent Space to the Pixel Space using a Variational Autoencoder (VAE)~\cite{kingma2013auto}.  The main contributor to inference latency is the forward propagation through $\epsilon_{\theta}$.
The architecture of $\epsilon_{\theta}$ is transitioning from U-Net~\cite{ronneberger2015u} to Diffusion Transformers (DiTs)~\cite{jiang2024megascale, chen2023pixart, ma2024latte}, driven by the scaling law that shows improved performance with increased model parameters and training data. Unlike U-Nets that use convolutional layers to capture spatial hierarchies, DiTs segment the input into latent patches and use the transformer's self-attention mechanism~\cite{vaswani2017attention} to model relationships within and across these patches. 
The input noisy latent representation is decomposed into patches, embedded into tokens, and fed into a series of DiT blocks, which generally include Multi-Head Self-Attention, Layer Norm, and Feedforward Networks. 
% Different DiT models may incorporate conditioning through methods like adaptive layer norm~\cite{peebles2023scalable}, cross-attention~\cite{chen2023pixart}, or extra input tokens~\cite{esser2024scaling}. As diffusion models handle higher-resolution images and longer sequences, they face a quadratic computational burden during inference, primarily due to the self-attention mechanism in DiTs.

\textbf{Input Temporal Redundancy:}
Diffusion Model entails the iterative prediction of noise from the input image or video. Recent research has highlighted the concept of \textit{input temporal redundancy}, which refers to the similarity observed in both the inputs and activations across successive diffusion timesteps~\cite{so2024frdiff, ma2024deepcache}. A recent study~\cite{ma2024learning} further investigates the distribution of this similarity across different layers and timesteps. Based on the redundancy, a branch of research caches the activation values and reuses them in the subsequent diffusion timesteps to prune computation. For example, in the context of the U-Net architecture, DeepCache updates the low-level features while reusing the high-level ones from cache~\cite{ma2024deepcache}. Additionally, \textsc{Tgate} caches the cross-attention output once it converges along the diffusion process~\cite{ma2024deepcache}. By contrast, in the domain of DiT models, $\Delta$-DiT proposes to cache the rear DiT blocks in the early sampling stages and the front DiT blocks in the later stages~\cite{chen2024delta}. PAB~\cite{zhao2024real} exploits the U-shaped attention pattern to mitigate temporal redundancy through a pyramid-style broadcasting approach. Finally, DiTFastAttn~\cite{yuan2024ditfastattn} identifies three types of redundancies, such as spatial redundancy, temporal redundancy, and conditional redundancy, and presents an attention compression method to speed up generation.

% \textbf{Parallel Inference for Diffusion Model}:
% Given the similar transformer architecture, tensor parallelism (TP)~\cite{shoeybi2019megatron} and sequence parallelism (SP)~\cite{jacobs2023deepspeed, liu2023ring, li2023lightseq}, which are commonly employed for efficient inference in LLMs, can be adapted for DiTs.
% TP partitions model parameters across multiple devices, enabling parallel computation and reducing memory requirements on individual devices. However, it necessitates an AllReduce operation for the outputs of both the Attention and Feedforward Network modules, resulting in communication overhead proportional to sequence length. For DiTs with exceptionally long sequences, this communication overhead becomes substantial. 
% In contrast, SP partitions the input image across multiple devices, utilizing methods such as Ulysses or Ring to communicate the Attention input and output tensors. 
% SP demonstrates superior communication efficiency compared to TP, but each device stores the entire model parameters, which can be memory intensive.
% DistriFusion~\cite{li2023distrifusion} leverages Input Temporal Redundancy to design an asynchronous sequence parallel method that utilizes stable activations.
% It is specifically designed for with a U-Net-based models.

\section{DiT Parallel Methods}
In this section, we delve into the application of tensor parallelism (TP), sequence parallelism (SP), and an asynchronous SP leveraging input temporal redundancy, namely DistriFusion, for parallel inference in DiTs. Following this, we introduce a novel patch-level pipeline parallel approach named PipeFusion that markedly improves communication efficiency compared to existing techniques.

In the remainder of this article, the following notations are used:
$p$ denotes the sequence length, which corresponds to the number of pixels in the latent space.
$hs$ represents the hidden size of the model.
$L$ stands for the number of network layers.
$N$ is the number of computing devices.

\subsection{Tensor Parallelism \& Sequence Parallelism}

Tensor parallelism (TP) partitions the linear layers of DiTs column-wise and row-wise. This approach requires two all-reduce operations per layer, incurring a communication overhead of $4O(p \times hs)$ per layer (Table~\ref{tab:parallel_schemes}). The communication cost scales linearly with the sequence length, leading to poor scalability, especially for large models such as the 0.6B-parameter PixArt model (Figure~\ref{fig:scale}). Additionally, TP poses implementation challenges for non-standard architectures like MM-DiT in Flux/SD3.

\begin{figure*}[htbp!]
\centering
\includegraphics[width=1\textwidth]{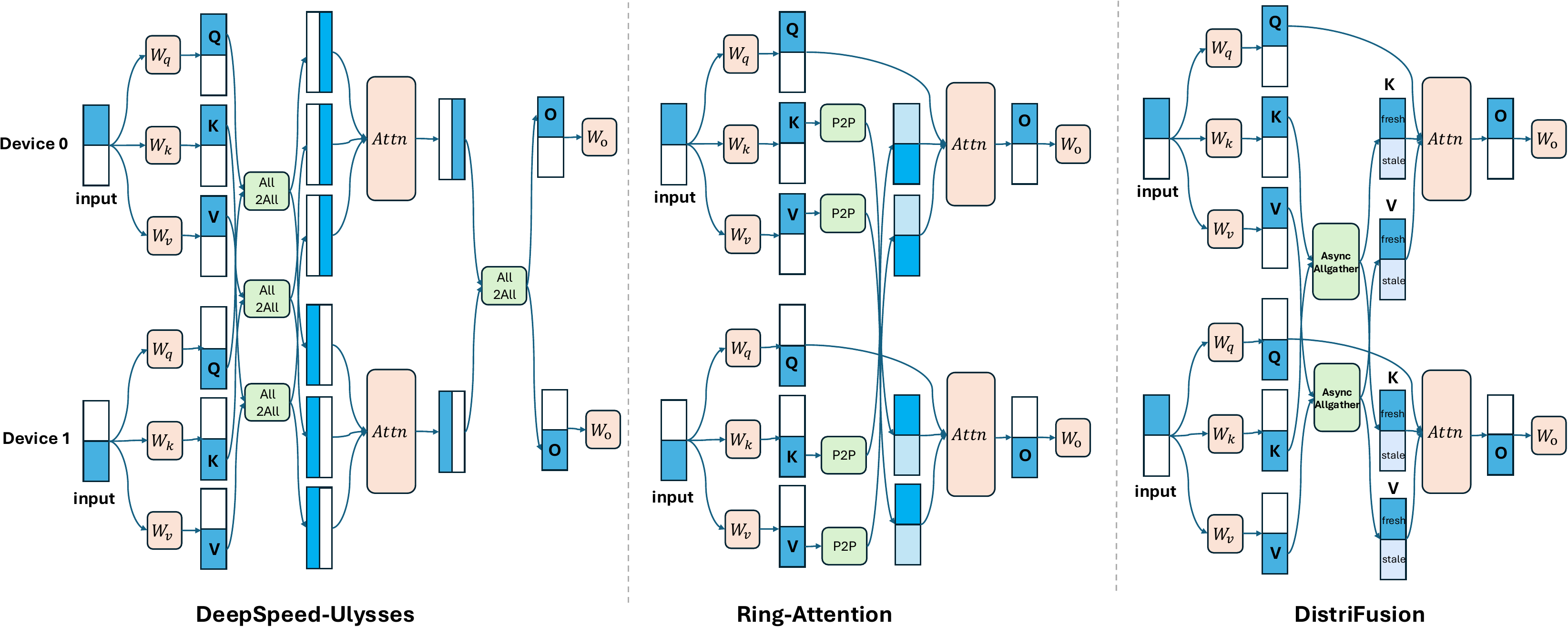}
\caption{Comparison of DistriFusion with Sequence Parallelism methods (DeepSpeed-Ulysses and Ring-Attention) for attention layers.}
\label{fig:seqparallel}
\end{figure*}

Sequence parallelism (SP) partitions images into patches, with each device processing a patch as input. Since attention mechanisms require global interactions, devices must communicate partial query (Q), key (K), and value (V) between each other. 
As shown in Figure~\ref{fig:seqparallel}, two notable SP implementations are DeepSpeed-Ulysses~\cite{jacobs2023deepspeed} (SP-Ulysses) and Ring-Attention~\cite{liu2023ring} (SP-Ring). SP-Ulysses uses All2All communication to switch between sequence and hidden-size partitioning during parallel attention, while SP-Ring implements distributed Flash Attention via point-to-point K/V subblock transfers.
It is worth noting that SP-Ulysses and SP-Ring can be combined to form USP (Unified Sequence Parallel)~\cite{fang2024unified}, which is often more efficient than using either method alone.

\subsection{DistriFusion: Asynchronous Sequence Parallelism}
DistriFusion is an asynchronous sequence parallelism method that leverages temporal redundancy in the input data. It uses slightly outdated (stale) activations from the previous timestep, rather than requiring fresh activations at each step. Despite this, DistriFusion maintains image generation accuracy without any perceptible loss in quality.

While DistriFusion was initially applied to U-Net-based SDXL models~\cite{podell2023sdxl}, it can be extended to DiT models as well. As illustrated in Figure~\ref{fig:seqparallel}, DistriFusion also operates by partitioning along the sequence dimension. The key distinction, however, lies in its use of asynchronous all-gather operations to collect K and V activations from remote devices. Due to the asynchronous nature of this communication, the activations for the current diffusion step are not immediately available; instead, they become accessible during the next diffusion step.

DistriFusion exploits this by utilizing a fraction $\frac{N-1}{N}$ of the K and V activations from timestep $T+1$, combined with a fraction $\frac{1}{N}$ of the local K and V from diffusion timestep $T$. This combination is used to compute the attention operation with the local queries at diffusion timestep $T$. The communication of K and V for timestep $T$ is intentionally overlapped with the network's forward computation at timestep $T$, thus hiding communication overhead during computation.

However, DistriFusion achieves this communication-computation overlap at the cost of increased memory usage. Each computational device is required to maintain communication buffers that store the complete spatial shape of the K and V activations, which amounts to $AL$ in total. Consequently, the memory cost of DistriFusion does not scale down with the addition of computational devices.

\subsection{PipeFusion: Patch-level Pipeline Parallel}
\label{sec:pipefuison}

% Existing parallel paradigms, whether Tensor Parallel or Sequence Parallel, require communication of activations for each DiTs block. 
% The sequence-level pipeline parallel method proposed by TeraPipe~\cite{li2021terapipe}, used in LLMs, can transmit only limited input Activations, and has been proven to be more suitable for long sequence input~\cite{qin2024mooncake}. 
% However, TeraPipe is designed for transformers using causal attention, where each token only attends to its previous tokens. In contrast, DiTs employ full attention, where each token attends to the computation of both previous and subsequent tokens.
% Nevertheless, by leveraging Input Temporal Redundancy, we found that TeraPipe can be perfectly applied to DiTs inference.
% We propose PipeFusion, a patch-level pipelined parallel approach for DiTs, which leverages the input temporal redundancy but exhibits enhanced communication and memory efficiency. 

\begin{figure*}[htbp]
\centering
\includegraphics[width=1\textwidth]{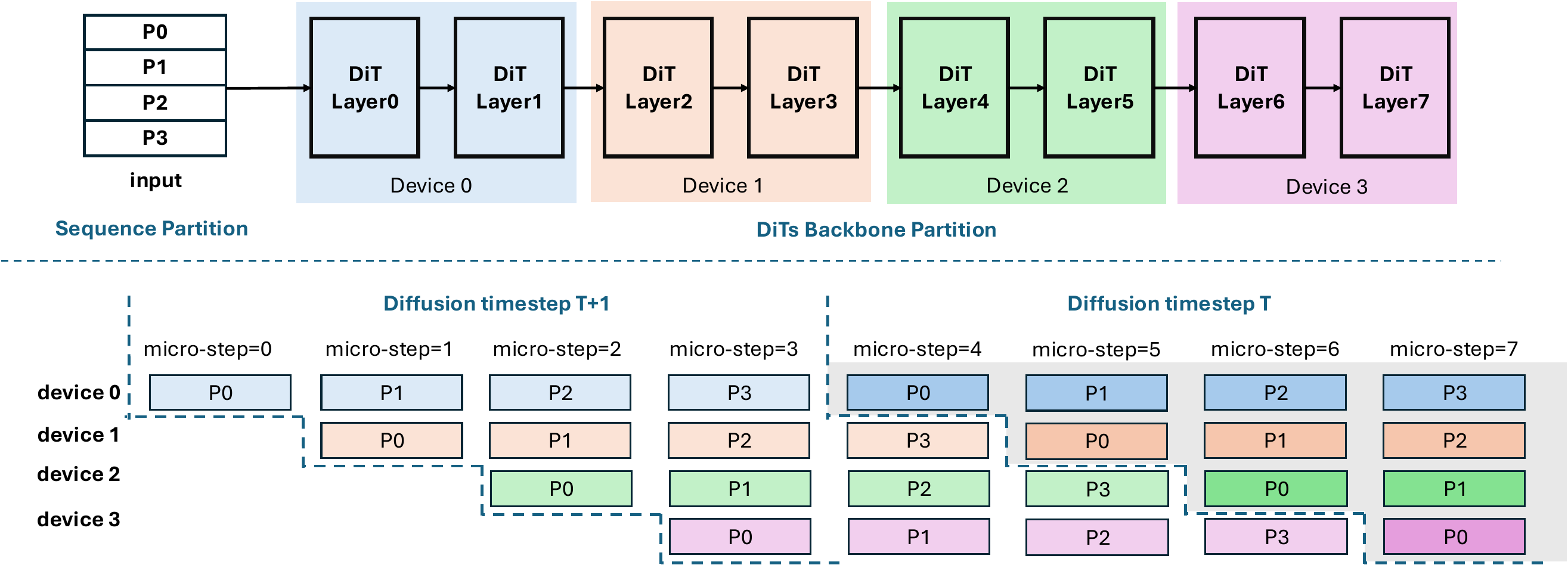}
\caption{Above: partitioning strategy for Input and DiTs backbone network. Below: Workflow of the PipeFusion as patch-level pipelined parallelism.}
\label{fig:pipefusion}
\end{figure*}

PipeFusion simultaneously partitions the input sequence and the layers of the DiTs backbone as shown at the top of Figure \ref{fig:pipefusion}.
PipeFusion partitions the DiTs model along the direction of data flow. 
Therefore, each partition contains a set of consecutive layers and is deployed on a GPU.
PipeFusion also partitions the input image into $M$ non-overlapping patches so that each GPU processes the computation for one patch with its assigned layers in parallel, and the diffusion routine runs efficiently in a pipelined manner. 
Pipelining requires synchronizations between devices, which would be efficient if the DiT workload is evenly distributed across each GPU. The even partition is easy to achieve as a DiT contains number of identical transformer blocks.

Suppose we are currently at diffusion timestep $T$, the previous timestep is $T+1$ as the diffusion process goes along the reversed order of timesteps.
The example in Figure~\ref{fig:pipefusion} demonstrates the pipeline workflow with $N = 4$ and $M = 4$, where the activation values of patches at timestep $T$ are highlighted.
Due to the input temporal redundancy, a device does not need to wait for the receiving of full spatial shape activations for the current timestep $T$ to start the computation of its own stage. 
Instead, it employs the stale activations from the previous timestep to provide context for the current computation. 
For instance, in micro-step 5, the activations for Patchs P0 and P1 are from timestep $T$, while others are from timestep $T + 1$.
In this way, each device maintains the full activations and there is no waiting time after the pipeline is initialized. Bubbles only exist at the beginning of the pipeline.
Suppose the total number of diffusion timesteps is $S$, then the effective computation ratio of the pipeline is $\frac{M \cdot S}{M \cdot S + N - 1}$. 
As $S$ is commonly chosen to be a large number for high-quality image generation, the effective computation ratio is high. For example, when $M=N=4$ and $S=50$, the ratio is equal to 98.5\%.
As long as the patch number \( M \) is greater than the parallel degree \( N \), there will be no bubble generation except for the initial startup overhead. We recommend setting \( M = N \). We have analyzed the impact of \( M \) on the final performance in Appendix Sec.~\ref{sec:ablstudy}.

Moreover, in PipeFusion, a device sends micro-step patch activations to the subsequent device via asynchronous P2P, enabling the overlap between communication and computation.
For example, at micro-step 4, device 0 receives P1's activation from diffusion timestep $T$ while concurrently computing the activation values for P0 at timestep $T$. Similarly, at the next micro-step, the transmission of P0 to device 1 can be hidden by the computation of P1 on the device.
% Since the sampler, i.e., DPM~\cite{lu2022dpm} or DDIM~\cite{song2020denoising}, are element-wise operations, all of the diffusion timesteps can be pipelined.

Similar to DistriFusion, before executing the pipeline, we usually conduct several diffusion iterations synchronously, called warmup steps.
During the warmup phase, patches are processed sequentially, resulting in low efficiency.
Though the warmup steps cannot be executed in a pipelined manner, the workflow is relatively small compared to the entire diffusion process, and thus, the impact on performance is negligible. 
We have analyzed the impact of warmup and propose the corresponding solutions in Appendix Sec.~\ref {sec:ablstudy}.

\begin{figure}[htbp]
\centering
\includegraphics[width=1\textwidth]{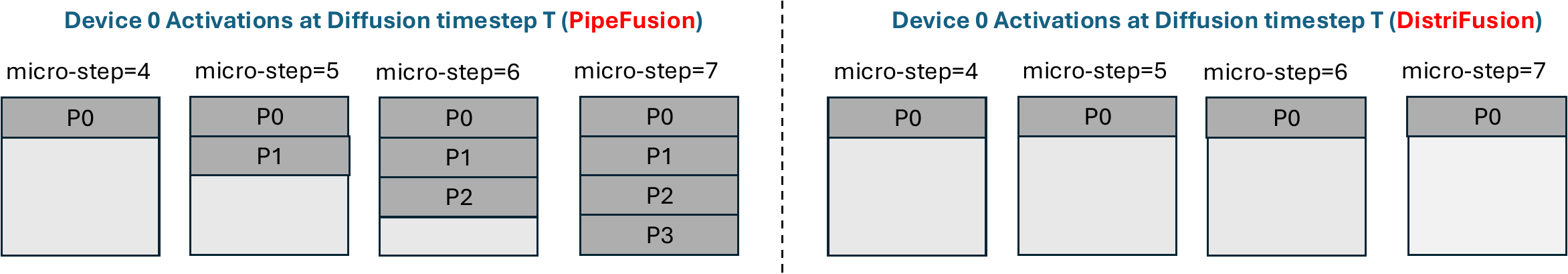}
\caption{The fresh part of activations during diffusion timestep $T$ of Figure~\ref{fig:pipefusion}. The dark gray represents fresh data and the light gray represents stable data. }
\label{fig:fresh_area}
\end{figure}

PipeFusion theoretically outperforms DistriFusion in terms of generation quality, considering the area of fresh activation.
As shown in Figure~\ref{fig:fresh_area}, within a single diffusion timestep, PipeFusion continuously increases the area of fresh activation as the pipeline micro-steps progress from diffusion timestep $4$ to $8$.
In contrast, throughout the entire diffusion process, DistriFusion constantly maintains one patch of fresh area out of the total $M$ patches.
Therefore, the results of PipeFusion are closer to the original diffusion method compared to DistriFusion.

The exploitation of input temporal redundancy requires a short warmup phase, since the behavior of diffusion synthesis changes qualitatively over the course of denoising. 
Accordingly, DistriFusion performs several warmup steps with standard synchronous patch parallelism before switching to its asynchronous mode. 
For PipeFusion, warmup steps cannot be pipelined and therefore introduce bubbles. 
However, because the number of warmup steps is typically very small compared to the total diffusion iterations, their overall impact on efficiency is limited. 
We analyze this effect in Sec.~\ref{sec:ablstudy}.

PipeFusion represents a fundamental departure from prior pipeline parallelization methods, such as GPipe~\cite{huang2019gpipe} and TeraPipe~\cite{li2021terapipe}.
Unlike GPipe, which splits along the batch dimension and depends on high concurrency across multiple requests, 
PipeFusion partitions along the sequence dimension to deliver efficient pipelined execution for a single image. 
Unlike TeraPipe, which is designed for causal attention in language models, 
PipeFusion is tailored to the full-attention mechanism of DiTs and exploits temporal redundancy to simultaneously improve efficiency and preserve generation quality.

\subsection{Comparison Between Different Parallelisms}
\label{sec:comparsion}

\begin{table}[h]
\centering
\caption{Comparison between different DiT parallel methods on a single diffusion timestep. Overlap denotes the overlap between communication and computation.}
\label{tab:parallel_schemes}
\begin{tabular}{cccccc}
\toprule
\multirow{2}{*}{\textbf{Method}}& \multicolumn{2}{c}{\textbf{Communication}} & \multicolumn{2}{c}{\textbf{Memory Cost}} \\
\cmidrule(lr){2-3} \cmidrule(lr){4-5}
& \textbf{Cost} & \textbf{Overlap} & \textbf{Model} & \textbf{KV Activations} \\
\midrule
Tensor Parallel & $4O(p \times hs)L$ & \xmark & $\frac{1}{N}P$ & $\frac{1}{N}KV$ \\

% DistriFusion(sync) & $O(p\times hs)L& \xmark & $P$ & 0 \\
% \hline
DistriFusion & $2O(p \times hs)L$ & \cmark & $P$ & $(KV)L$ \\

SP-Ring & $2O(p \times hs)L$ & \cmark & $P$ & $\frac{1}{N}KV$ \\

SP-Ulysses & $\frac{4}{N}O(p \times hs)L$ & \xmark & $P$ & $\frac{1}{N}KV$ \\

PipeFusion & 2$O(p \times hs)$ & \cmark & $\frac{1}{N}P$ & $\frac{1}{N}(KV)L$ \\
\bottomrule
\end{tabular}
\end{table}

Compared with existing parallel solutions, PipeFusion demonstrates superior efficiency in communication and memory usage.
Firstly, PipeFusion transmits between computational devices only the activations that serve as inputs and outputs for a series of consecutive transformer layers belonging to a stage, significantly reducing the communication bandwidth requirement in comparison to other methods that transmit the KV activations for all the $L$ layers.
In particular, the total communication cost for PipeFusion is $2O(p\times hs)$, not associated with the number of layers $L$.
Secondly, each device in PipeFusion stores only $\frac{1}{N}$ of the parameters belonging to its specific layers.
Since the usage of stale KV for attention computation requires each device to maintain the full spatial KV for the $\frac{L}{N}$ layers on the device, the overhead is significantly smaller than that of DistriFusion and diminishes as the number of devices increases. 

% It should be emphasized that the precision preservation design of DistriFusion for the group normalization layers in U-Net, specifically the \textit{Corrected Asynchronous GroupNorm}, can be seamlessly applied in PipeFusion, although DiTs do not employ a group norm layer.

% \begin{figure}[htbp]
% \centering
% \includegraphics[width=0.8\textwidth]{./figures/convergence.pdf}
% \caption{The fresh part of activations during diffusion timestep $T$. The dark gray represents fresh data and the light gray represents stable data. }
% \label{fig:fresh_area}
% \end{figure}

In the following, we summarize the theoretical cost of existing methods in Table~\ref{tab:parallel_schemes}.
The communication cost is calculated by the product of the number of elements to transfer with an \textit{algorithm bandwidth (algobw)} factor related to communication type~\footnote{https://github.com/NVIDIA/nccl-tests/blob/master/doc/PERFORMANCE.md}.
For collective algorithms of AllReduce, AllGather and AllToAll, the corresponding algobw factors are \( 2\frac{n-1}{n} \), \( \frac{n-1}{n} \), and 1. 
In the table, we approximate the term $O(\frac{n-1}{n})$ to $O(1)$ for simplicity.

Among all methods, PipeFusion has the lowest communication cost, as long as $N<2L$, which is easy to satisfy as the number of network layers $L$ is typically quite large, e.g., $L=38$ in Stable-Diffusion-3.
In addition, PipeFusion overlaps communication and computation.
Ulysses Seq Parallel exhibits a decreasing communication cost with increasing $N$, outperforming the remaining three methods. 
However, its communication cannot be hidden by computation.
Ring Seq Parallel and DistriFusion have similar communication costs and overlapping behaviors. 
The distinction lies in the scope of overlapping: computation in Ring Seq Parallel overlaps within the attention module, whereas that in DistriFusion overlaps throughout the entire forward pass.

To analyze the memory costs, we denote $P$ as the total number of model parameters.
In PipeFusion and tensor parallelism, the memory cost decreases as more GPUs are utilized, which is a nice property with the rapid growth of the DiT model size.
Both PipeFusion and DistriFusion maintain a KV buffer for each transformer layer leading to significant activation memory overhead, especially for long sequences.
The KV buffer in PipeFusion decreases as the number of devices $N$ increases, whereas DistriFusion does not exhibit such a reduction.

\section{Experiments}

In our experimental setup, we deployed our trials on an 8$\times$L40-48GB (PCIe Gen4x16) cluster to evaluate three prominent DiT models: \textbf{Pixart}~\cite{chen2023pixart, chen2024pixart}: Its backbone is a transform model that encompasses 0.6B (billion) parameters. Pixart draws its architectural foundation from the original DiTs framework, with a pivotal enhancement: the integration of cross-attention modules designed to incorporate text conditions into the model's processing. \textbf{Stable-Diffusion-3}~\cite{esser2024scaling} (SD3-medium): which leverages a Multimodal (MM)-DiT and its backbone transformer model contains 2B parameters. As the largest open-source variant available, SD3-medium employs a 20-step FlowMatchEulerDiscreteScheduler. \textbf{Flux.1-dev}~\cite{blackforestlabs2023announcement}: Its backbone transformer model contains 12B parameters and features an enhanced MM-DiT design. Flux.1-dev also relies on a 28-step FlowMatchEulerDiscreteScheduler.

% The inputs to MM-DiT blocks are the sequence concating the embeddings of the text and image inputs, and use two separate sets of weights for the two modalities. 

% \begin{itemize}[% 设置itemize环境的缩进
%   leftmargin=*, % 左边距，*表示与当前段落的左边距相同
%   itemindent=0pt % 项目标识的缩进
% ]
%     \item Tensor Parallelism: It splits parameter tensors across GPUs referring to Megatron-LM, which applies two allreduce communications for a DiT block.
%     \item Sequence Parallelism: The DeepSpeed-Ulysses and Ring-Attention is applied on the attention module on DiT.
%     \item  DistriFusion: We adapt the U-Net implementation to DiT.
%     \item Original: a serial implementation on 1 GPU using huggingface acceleration.
% \end{itemize}

\subsection{Performance Results}
This section compares the performance of different parallel methods on 8$\times$L40 GPUs, connected by PCIe Gen4.
This section only presents the latency of the DiT except for the latency of the VAE decoding, since VAE module is the same across different parallel methods.
Since the Pixart and SD3 can utilize Classifier-Free Guidance (CFG) ~\cite{ho2022classifier}, we additionally implemented the CFG parallel.
For an input prompt, CFG parallel individually computes the forward tasks for both unconditional guidance and text guidance. 
It leverages inter-image parallel to perform these two tasks, collecting the results after each diffusion step is completed.
For PipeFusion, we select the best latency performance by searching the patch number $M$ from 2, 4, 8, 16, 32.
All figures show the average latency of 5 runs.
We employ a default warmup step of 1 for both DistriFusion and PipeFusion. The software stack utilized includes PyTorch 2.4.1, CUDA Runtime 12.1.105, and diffusers 0.30.3.

\subsubsection{Latency Evaluation}
\begin{figure*}[htbp!]
\centering
\includegraphics[width=1\textwidth]{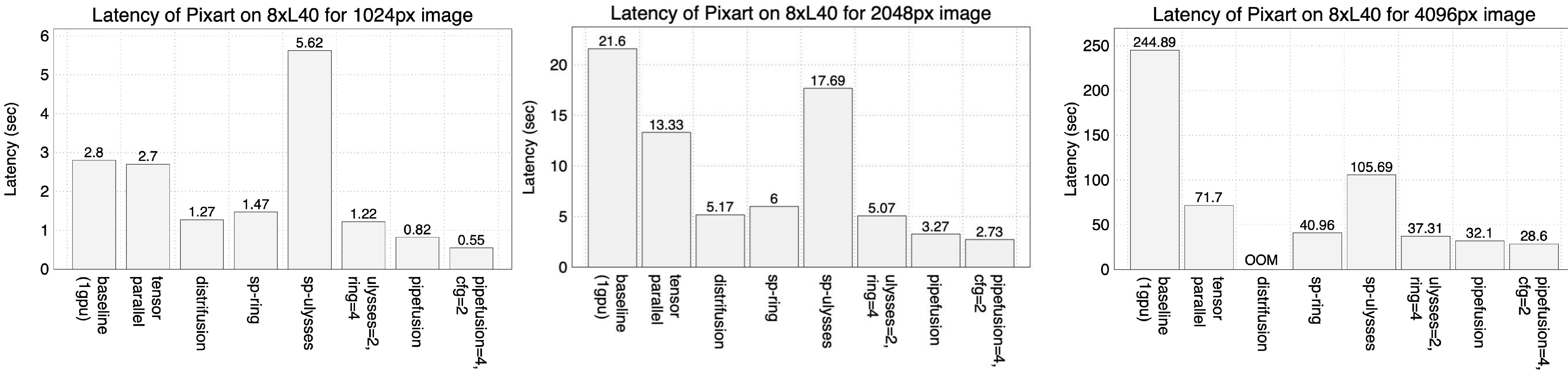}
\caption{Latency on Pixart of various parallel approaches on two image generation tasks with the 20-Step DPMSolverMultistepScheduler.}
\label{fig:pixart}
\end{figure*}

\textbf{Pixart:} Figure~\ref{fig:pixart} shows the inference latency of Pixart on image generation tasks with resolution as 1024px (1024$\times$1024), 2048px (2048$\times$2048) and 4096px (4096$\times$4096) on 8$\times$L40.

% \begin{figure}[htbp!]
%   \centering
%   \begin{subfigure}[b]{0.48\textwidth}
%     \centering
%     \includegraphics[width=\textwidth]{./figures/iclr/sd3-12k-crop.pdf}
%     \caption{Latency on SD3-medium of various parallel approaches on two image generation tasks with the 20-Step FlowMatchEulerDiscrete Scheduler.}
%     \label{fig:sd3}
%   \end{subfigure}
%   \hfill
%   \begin{subfigure}[b]{0.48\textwidth}
%     \centering
%     \includegraphics[width=\textwidth]{./figures/iclr/flux-12k-crop.pdf}
%     \caption{Latency on Flux.1-dev of various parallel approaches on three image generation tasks with the 28-Step FlowMatchEulerDiscrete Scheduler.}
%     \label{fig:flux}
%   \end{subfigure}
%   \caption{Comparison of latency between SD3-medium and Flux.1-dev under different parallel approaches.}
%   \label{fig:parallel_latency}
% \end{figure}

\begin{figure}[htbp!]
  \centering
  \includegraphics[width=0.7\textwidth]{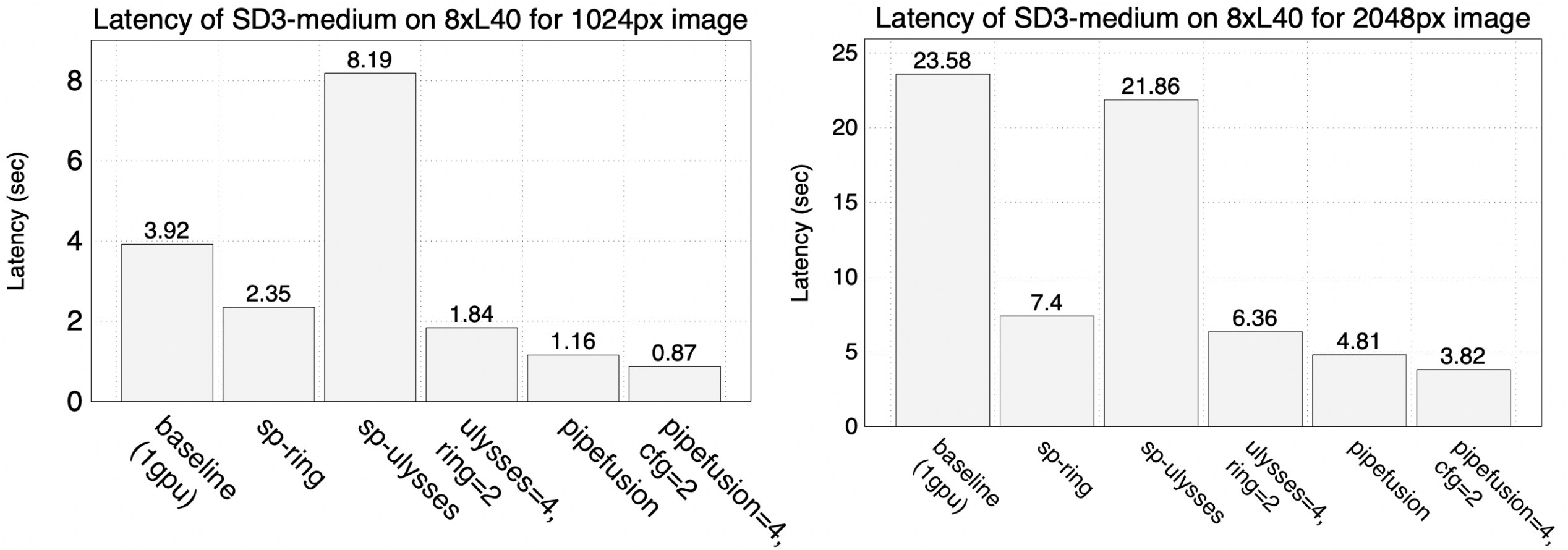}
  \caption{Latency on SD3-medium of various parallel approaches on two image generation tasks with the 20-Step FlowMatchEulerDiscrete Scheduler.}
  \label{fig:sd3}
\end{figure}

\begin{figure}[htbp!]
  \centering
  \includegraphics[width=0.7\textwidth]{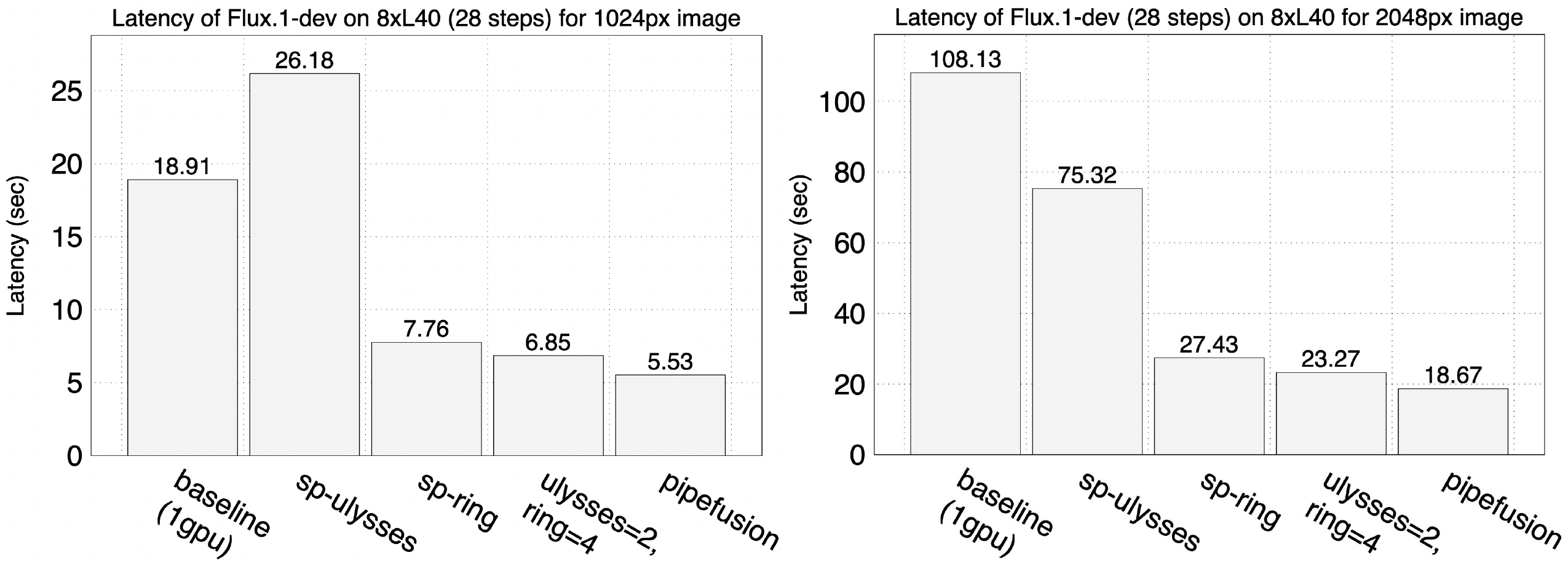}
  \caption{Latency on Flux.1-dev of various parallel approaches on three image generation tasks with the 28-Step FlowMatchEulerDiscrete Scheduler.}
  \label{fig:flux}
\end{figure}

PipeFusion exhibits the lowest latency among parallel methods across all three tasks. 
\textbf{\textit{PipeFusion is 1.48$\times$, 1.55$\times$, and 1.16$\times$ faster than the second best parallel approaches on the 1024px, 2048px and 4096px tasks.}}

The highest latency parallelism is always SP-Ulysses, primarily due to the bandwidth bottleneck between CPU sockets by conducting All2All on 8 GPUs via PCIe.
SP-Ring exhibits significantly lower latency compared to SP-Ulysses.
An efficiency method USP~\cite{fang2024unified} can further improve SP performance by hybridizing SP-Ulysses and SP-Ring.
It views the process group as a 2D mesh where the columns are SP-Ring groups and the rows are SP-Ulysses groups.
USP parallelizes the attention head dimension in SP-Ulysses groups and parallelizes the sequence dimension in SP-Ring groups. 
USP yields a lower latency than SP-Ring, as illustrated by the ulysses=2 and ring=4 columns in the figure.

Additionally, tensor parallel also shows high latency, which aligns with our analysis in section~\ref{sec:comparsion}.
DistriFusion exhibits slightly higher latency compared to the USP (ulysses=2 and ring=4), and it encounters out-of-memory (OOM) issues on the 4096px task.

Using the CFG parallel, the 8 GPUs are divided into two groups, each performing PipeFusion independently, which can further reduce latency. This approach (cfg=2, pipefusion=4 in the figure) achieves a latency reduction of 5.09x, 7.91x, and 8.59x compared to the baseline (1 GPU).
PipeFusion achieves perfect scalability in tasks 2048px and 4096px.

% \begin{figure}[htbp!]
% \centering
% \includegraphics[width=0.5\textwidth]{./figures/iclr/flux-12k-crop.pdf}
% \caption{Latency on Flux.1-dev of various parallel approaches on three image generation tasks with the 28-Step FlowMatchEulerDiscrete Scheduler.}
% \label{fig:flux}
% \end{figure}

\begin{figure*}[htbp!]
\centering
\includegraphics[width=1\textwidth]{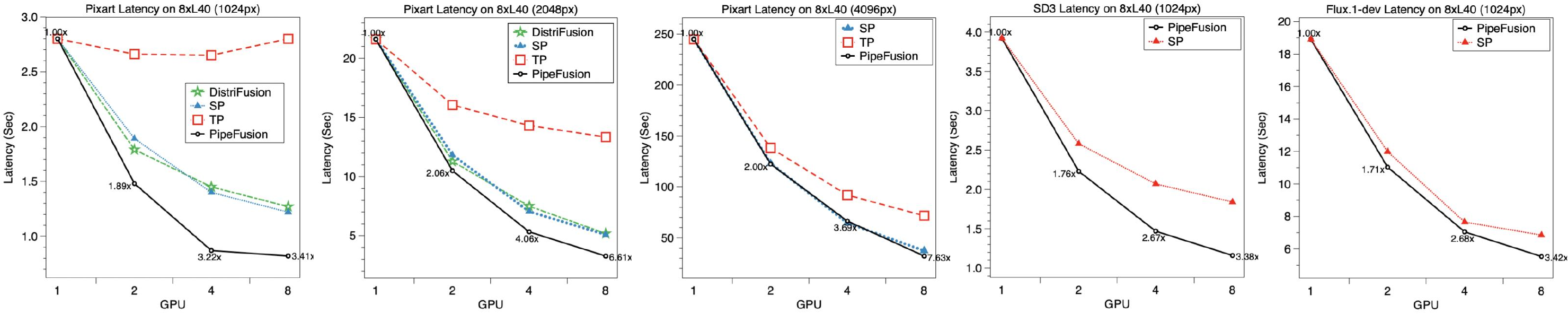}
\caption{Scalability Analysis of Three DiTs on 8$\times$L40: Leftmost figures depict Pixart performance on 1024px, 2048px, and 4096px images. The second from the right shows SD3 performance on 1024px. The rightmost figure illustrates Flux.1-dev performance on 1024px. The speedup to 1 GPU baseline is labeled on the points of pipefusion curves.}
\label{fig:scale}
\end{figure*}

\textbf{Stable Diffusion 3:}
Figure~\ref{fig:sd3} shows the inference latency of SD3-medium on image generation tasks with resolution as 1024px, 2048px on 8$\times$L40.
The model is unable to generate 4096px images due to positional encoding limitations.
Since MM-DiT applied in SD3 is a non-standard transformer layer, arranging column-wise and row-wise weight partitioning poses significant challenges, and we did not implement tensor parallelism.
Additionally, due to potential OOM issues with DistriFusion and inferior performance compared with USP, we did not implement it for SD3.

PipeFusion outperforms the best sequence parallelism in these two cases, both of which are also the USP with ulysses=4, ring=2, \textbf{\textit{achieving speedups of 1.57$\times$ and 1.30$\times$.}}
When combined with CFG parallel, the pipefusion=4, cfg=2 scheme on 8 GPUs achieves speedups of 3.11$\times$ and 8.16$\times$ compared to the baseline (1 GPU).

\textbf{Flux.1-dev:}
Figure~\ref{fig:flux} illustrates the inference latency of SD3-medium on image generation tasks with resolutions of 1024px and 2048px, respectively, using 8$\times$L40.
Flux.1 employs a similar MM-DiT architecture to SD3 but extends the network depth to 57 layers.

In both scenarios, PipeFusion surpasses the best sequence parallelism, which is also the USP with ulysses=4 and ring=2, \textit{\textbf{achieving speedups of 1.23$\times$ and 1.25$\times$}}.
PipeFusion offers 3.42$\times$ and 5.79$\times$ speedup compared to the baseline (1 GPU) for the two cases.

\textbf{Scalability}: 
We evaluated the scalability of inference for three models by scaling from 1 to 2, 4, and 8 GPUs. Figure~\ref{fig:scale} illustrates the scalability of different parallel approaches. 
PipeFusion and DistriFusion both apply a single warmup iteration, while for SP we adopt the USP, which searches the optimal combination of Ulysses and Ring degrees. The results show that PipeFusion consistently achieves the best scalability across all five scenarios. Given that PixArt has the smallest model size, tensor parallelization unsurprisingly exhibits the worst scalability.  

At 1024px and 2048px resolutions, PipeFusion substantially outperforms both SP and DistriFusion. Even at 4096px, where DistriFusion encounters out-of-memory (OOM) issues and thus provides no valid data, PipeFusion continues to deliver strong performance: on 8$\times$L40 GPUs, it achieves 32.1s end-to-end latency versus SP’s best 37.3s, corresponding to a 1.16$\times$ speedup. While this speedup appears smaller than at lower resolutions, PipeFusion’s communication efficiency remains a clear advantage. Specifically, PipeFusion reduces communication share to only 4.6\% of total latency at 4096px, compared with 17.9\% for SP. 

Here, the communication share is defined as

\begin{equation}
\text{CommShare} = \frac{T_{\text{E2E},\,8\text{GPU}} - \tfrac{T_{\text{single}}}{8}}{T_{\text{E2E},\,8\text{GPU}}},
\end{equation}
which subtracts 1/8 of the single-GPU latency from the observed 8-GPU latency. For example,
\[
\frac{32.1 - 244.89/8}{32.1} \approx 4.6\% \quad \text{(PipeFusion)}, \qquad
\frac{37.3 - 244.89/8}{37.3} \approx 17.9\% \quad \text{(SP)}.
\]

In absolute terms, SP’s communication cost is 6.69s, whereas PipeFusion lowers it to 1.49s --- a 78\% reduction. This shows that even in the compute-bound regime, PipeFusion robustly suppresses communication overhead.

% \textbf{Scalability}:
% We evaluated the scalability of inference for three models by scaling from 1 to 2, 4, and 8 GPUs. Figure~\ref{fig:scale} illustrates the scalability of different parallel approaches.
% PipeFusion and DistriFusion both apply a single warmup iteration.
% We applied USP to represent SP by searching for the optimal combination of ulysses degree and ring degree to achieve the best performance.
% The results show that PipeFusion consistently exhibits the best scalability across all five scenarios. Given that Pixart has the smallest model size, tensor parallelization unsurprisingly exhibits the worst scalability.
% Except for the Pixart 4096px case, PipeFusion significantly outperforms both SP and DistriFusion. In the Pixart 4096px scenario, PipeFusion and SP exhibit similar scalability, while DistriFusion encounters out-of-memory (OOM) issues and thus lacks corresponding data. This is likely due to the exceptionally long sequence, where the communication overhead becomes less significant, thereby diminishing the advantage of PipeFusion's communication efficiency.

\subsubsection{Memory Efficiency}
We collect the memory usage excluding the VAE usage, which involves the computation of the text-encoder and DiTs backbones, as shown inAs shown in Figure~\ref{fig:ditinfer}.
The figure includes memory usage for "parameters", which encompasses both the text encoder and transformers, as well as for "activations", which includes activations and temporary buffers.
Note that SP-Ulysses, SP-Ring and USP exhibit similar memory consumption, denoted as SP in the figure.

\begin{figure}[htbp!]
  \centering
  \begin{subfigure}[b]{0.43\textwidth}
    \centering
    \includegraphics[width=\textwidth]{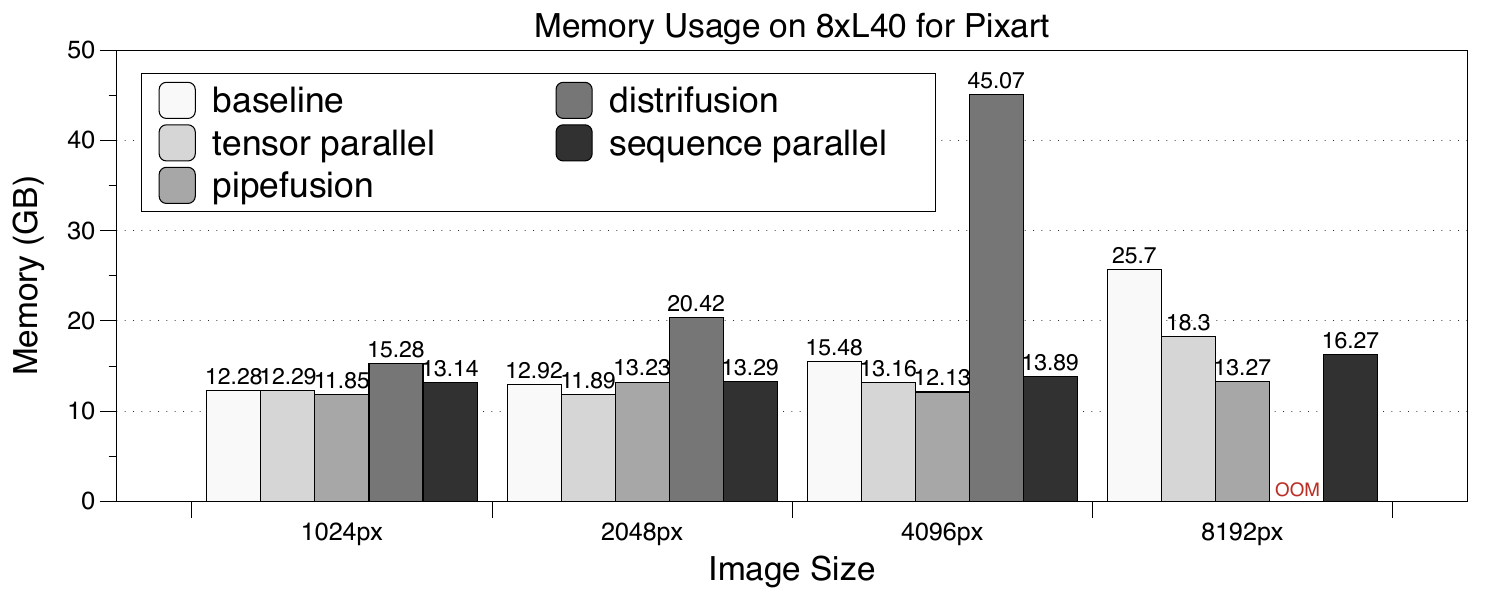}
    \caption{Max GPU Memory Usage of Various Approaches for Pixart Image Generation Tasks Across Four Resolutions.}
    \label{fig:pixart-memory-A100}
  \end{subfigure}
  \hfill
  \begin{subfigure}[b]{0.56\textwidth}
    \centering
    \includegraphics[width=\textwidth]{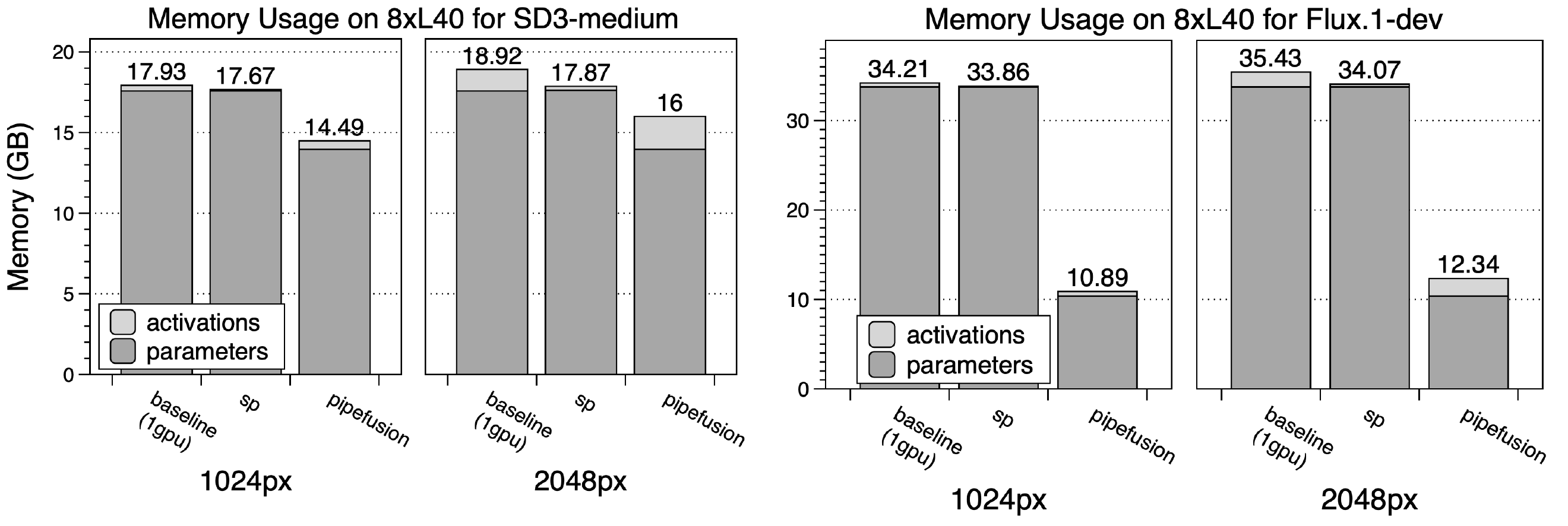}
    \caption{Max GPU Memory Usage on 8 GPUs for SD3-medium and Flux.1 on 1024px and 2048px image generation tasks.}
    \label{fig:sd3-flux-memory}
  \end{subfigure}
  \caption{Comparison of GPU memory usage across different models and resolutions.}
  \label{fig:memory-comparison}
\end{figure}

The maximum memory usage of different parallel approaches for Pixart is depicted in Figure~\ref{fig:pixart-memory-A100}. 
The Pixart model consists of a 0.6B parameter DiT backbone (2.3 GB on disk) and T5-based text encoders (18GB on disk).
So the memory consumption of parameters is dominated by the text encoder. 
DistriFusion maintains full spatial shape K, V tensors for each layer, resulting in an increase in memory consumption as the image resolutions grow. 
In contrast, PipeFusion only needs to store 1/$N$ of the full spatial shape K, V tensors. 
Consequently, on 8$\times$GPUs, its memory footprint is comparable to that of Tensor Parallel and Sequence Parallel. 
Remarkably, even for tasks with an 8192px resolution, PipeFusion exhibits the lowest memory consumption. 

% \begin{figure}[htbp!]
%   \centering
%   \includegraphics[width=0.6\textwidth]{./figures/iclr/Pixart-memory.pdf}
%   \caption{Max GPU Memory Usage of Various Approaches for Pixart Image Generation Tasks Across Four Resolutions.}
%   \label{fig:pixart-memory-A100}
% \end{figure}

% \begin{figure}[htbp!]
%   \centering
%   \includegraphics[width=0.7\textwidth]{./figures/iclr/memory_sd3_flux-crop.pdf}
%   \caption{Max GPU Memory Usage on 8 GPUs for SD3-medium and Flux.1 on 1024px and 2048px image generation tasks.}
%   \label{fig:sd3-flux-memory}
% \end{figure}

The maximum memory usage of sequence parallel and PipeFusion for SD3-medium and Flux.1-dev is depicted in Figure~\ref{fig:sd3-flux-memory}. 
The SD3-medium consists of a 2B parameter DiTs backbone (7.8 GB on disk) and text encoders of nearly 19GB on disk.
The Flux.1-dev consists of a 12B parameter DiTs backbone (23 GB on disk) and text encoders of nearly 9.1GB on disk.
The memory cost of PipeFusion is also significantly less than that of sequence parallelism.
For the 12B Flux.1-dev, PipeFusion significantly reduces the memory footprint of model parameters. However, PipeFusion increases the memory usage of activations due to the consumption of the KV Buffer. Nevertheless, this increase is relatively minor compared to the overall model parameter memory usage.
\textbf{\textit{The overall memory usage of PipeFusion is 32\% and 36\% of SP on 1024px and 2048px cases using Flux.1}}.

\subsection{Quality Results}
Table~\ref{tab:fid_scores} presents the Fréchet Inception Distance (FID)~\cite{heusel2017gans} scores for PipeFusion, DistriFusion, and the Original model, evaluated on the Pixart and Flux.1 datasets. A lower FID score is indicative of superior performance in terms of image quality. The results clearly demonstrate that PipeFusion consistently outperforms DistriFusion across identical device configurations, achieving significantly lower FID scores. This finding is in direct alignment with the theoretical analysis presented in Sec.~\ref{sec:pipefuison}, where PipeFusion is shown to leverage fresher activations while effectively reducing the impact of stale ones, thereby enhancing overall performance.

Additionally, we provide a comprehensive visual comparison in Fig.~\ref{fig:quality} of Appendix~\ref{sec:app_image_quality}, where image samples generated by PipeFusion, DistriFusion, and the Original model are displayed. The images generated by PipeFusion are nearly indistinguishable from the original images to the human eye.

\begin{table*}[ht!]
\small
    \centering
    \begin{tabular}{ccccccccccc}
        \toprule
        \multicolumn{2}{c}{\textbf{Model}} & \multicolumn{6}{c}{\textbf{PixArt-XL-2-256-MS}} & \multicolumn{3}{c}{\textbf{FLUX.1[dev]}} \\ 
        \midrule
        \multicolumn{2}{c}{\textbf{Method}} & Original & \multicolumn{2}{c}{DistriFusion} & \multicolumn{3}{c}{PipeFusion} & Original & \multicolumn{2}{c}{PipeFusion} \\ % \hline
        \cmidrule(lr){3-3}\cmidrule(lr){4-5}\cmidrule(lr){6-8}\cmidrule(lr){9-9}\cmidrule{10-11} \multicolumn{2}{c}{\textbf{Device Number}} & 1 & 4 & 8 & 2 & 4 & 8 & 1 & 4 & 8 \\ % \hline
        \midrule
        \multirow{2}{*}{\textbf{FID $\downarrow$}}
        & \textbf{w/ G.T.} & 22.78 & 30.69 & 41.81 & 23.60 & 25.23 & 28.46 & 25.03 & 24.17 & 25.97 \\ % \hline
        & \textbf{w/ Orig.} & - & 8.41 & 19.81 & 1.67 & 3.11 & 6.09 & - & 4.01 & 5.93 \\ % \hline
        \bottomrule
    \end{tabular}
    \caption{FID Scores for Parallel Methods on the Pixart and Flux.1. w/ G.T. means calculating the FID metrics with the ground-truth images. w/ Orig. means calculating the FID metrics with the images generated by the 1-Device original implementation.}
    \label{tab:fid_scores}
\end{table*}

% \begin{table}[htbp!]
%     \scriptsize
%     \centering
%     \begin{tabular}{cccccccccccc}
%     \toprule
%     \multicolumn{2}{c}{\textbf{Model}} & \multicolumn{6}{c}{\textbf{PixArt-XL-2-256-MS}} & \multicolumn{4}{c}{\textbf{FLUX.1[dev]}} \\ 
%     \midrule
%     \multicolumn{2}{c}{\textbf{Method}} & Original & \multicolumn{2}{c}{DistriFusion} & \multicolumn{3}{c}{PipeFusion} & Original & \multicolumn{3}{c}{PipeFusion} \\ % \hline
%     \cmidrule(lr){3-3}\cmidrule(lr){4-5}\cmidrule(lr){6-8}\cmidrule(lr){9-9}\cmidrule{10-12} 
%     \multicolumn{2}{c}{\textbf{Device Number}} & 1 & 4 & 8 & 2 & 4 & 8 & 1 & 2 & 4 & 8 \\ % \hline
%     \midrule
%     \multirow{2}{*}{\textbf{FID $\downarrow$}}
%     & \textbf{w/ G.T.} & 22.78 & 30.69 & 41.81 & 23.60 & 25.23 & 28.46 & 23.52 & 23.30 & 24.17 & 25.97 \\ % \hline
%     & \textbf{w/ Orig.} & - & 8.41 & 19.81 & 1.67 & 3.11 & 6.09 & - & 2.22 & 3.06 & 4.64 \\ % \hline
%     \bottomrule
%     \end{tabular}
%     \caption{FID Scores for Parallel Methods on the Pixart and Flux.1 w/ G.T. means calculating the FID metrics with the ground-truth images. w/ Orig. means calculating the FID metrics with the images generated by the 1-Device original implementation.}
%     \label{tab:fid_scores}
% \end{table}

% \begin{table}[h]
%   \centering
%   \caption{Experimental Device}
%   \label{tab:equipment}
%   \begin{tabular}{|c|c|c|}
%     \hline
%     \textbf{Name} & \textbf{p2p Bandwidth} & \textbf{Memory} \\
%     \hline
%     A100(PCIe) &  23 GB/s & 80 GB \\
%     A100(NVLink) & 268 GB/s & 40 GB \\
%     L20(PCIe) & 26 GB/s & 40 GB\\
%     \hline
%   \end{tabular}
% \end{table}

\section{Limitations \& Discussion}
While this paper has demonstrated PipeFusion’s advantages over tensor parallelism and sequence parallelism, we acknowledge that it is not a universal solution. 
In fact, combining PipeFusion with existing approaches into a hybrid parallelization scheme is likely the most practical direction for deployment. 
Such a design enables efficient scaling of DiT inference across large cross-node configurations: sequence parallelism can be exploited within NVLink-connected, high-bandwidth settings, while PipeFusion is particularly effective across inter-node, low-bandwidth environments.

A key limitation of PipeFusion lies in its dependence on multiple diffusion steps. 
For models with only a handful of steps—such as distilled variants like the 4-step Flux.1-schnell~\cite{flux1_schnell} or one-step diffusion methods~\cite{yin2024one,geng2025mean}—temporal redundancy is minimal, making PipeFusion less effective. 
However, this does not pose a major constraint in practice, as state-of-the-art DiT models generally require a substantial number of steps to deliver competitive quality.

Another promising extension is the application of PipeFusion to video generation. 
Architecturally, PipeFusion is already compatible with multimodal DiTs (MM-DiTs) such as Flux.1 and Stable Diffusion~3, where cross-modal conditions are injected through unified attention. 
Recent video models (e.g., Wan2.1~\cite{wan2025wan}, HunyuanVideo~\cite{kong2024hunyuanvideo}, CogVideo~\cite{hong2022cogvideo}) adopt similar MM-DiT backbones, suggesting that PipeFusion can be applied with minimal adaptation. 
From the tensor-shape perspective, both image and video DiTs operate on latent tensors of shape $(B, L, H)$, with video models flattening spatial and temporal dimensions into a unified sequence dimension. 
Because PipeFusion partitions along the sequence dimension, this structural difference does not affect applicability. 
From the performance perspective, video models are typically larger and involve longer sequence lengths, which amplifies communication overhead for SP. 
PipeFusion mitigates this overhead through parameter partitioning and pipelined execution, offering reduced memory footprint and stronger scalability. 
This suggests that video generation is not only a natural but also a highly favorable domain for applying PipeFusion.

\section{Conclusion}
% This paper introduces PipeFusion, an innovative parallel methodology for accelerating the inference of Diffusion Transformers (DiTs) on multiple devices. By exploiting the high similarity between inputs from successive diffusion steps and employing a patch-level pipelined approach, PipeFusion significantly reduces both communication bandwidth and memory demands.
% Experimental results demonstrate that PipeFusion achieves state-of-the-art performance on 8×L40 PCIe-connected GPUs for Pixart, Stable-Diffusion 3, and Flux.1 models. For Pixart, PipeFusion achieves a 1.16× to 1.55× speedup over the sequence parallel approaches. For SD3, PipeFusion achieves a 1.30× to 1.57× speedup over sequence parallelism methods. For Flux.1-dev, PipeFusion achieves a 1.23× to 1.25× speedup over sequence parallelism methods.

We introduced PipeFusion, a patch-level pipeline parallelism strategy tailored for Diffusion Transformers. By reusing temporally redundant activations and overlapping communication with computation, PipeFusion reduces both bandwidth demands and memory footprint while preserving image quality. Experiments on Pixart, Stable Diffusion 3, and Flux.1 demonstrate consistent latency improvements over strong baselines, achieving up to 1.5× faster inference on commodity multi-GPU clusters.

Beyond efficiency gains, PipeFusion highlights a broader systems insight: temporal redundancy in diffusion processes can be systematically exploited to design new forms of parallelism. While our study focused on single-node, multi-GPU inference, the approach naturally composes with existing sequence or tensor parallelism, offering a path toward scalable deployment on large clusters. We hope this work inspires further exploration of hybrid parallel strategies that close the gap between rapidly growing generative models and the practical requirements of serving them at scale.

% \section{References}

%%%%%%%%%%%%%%%%%%%%%%%%%%%%%%%%%%%%%%%%%%%%%%%%%%%%%%%%%%%%

\bibliographystyle{unsrt}  
\bibliography{ref}  %%% Remove comment to use the external .bib file (using bibtex).

\section*{NeurIPS Paper Checklist}

\begin{enumerate}

\item {\bf Claims}
    \item[] Question: Do the main claims made in the abstract and introduction accurately reflect the paper's contributions and scope?
    \item[] Answer: \answerYes{} % Replace by \answerYes{}, \answerNo{}, or \answerNA{}.
    \item[] Justification: The paper clearly states its contribution of proposing PipeFusion, a novel patch-level pipeline parallelism method for efficient inference of diffusion transformer models. It highlights the key ideas of leveraging input temporal redundancy and reducing communication and memory costs compared to existing parallel methods. The claims are supported by the experimental results demonstrating state-of-the-art performance on multiple models and configurations.

\item {\bf Limitations}
    \item[] Question: Does the paper discuss the limitations of the work performed by the authors?
    \item[] Answer: \answerYes{} % Replace by \answerYes{}, \answerNo{}, or \answerNA{}.
    \item[] Justification: The paper discusses limitations in Section 5, noting that PipeFusion is not well-suited for models with a limited number of diffusion steps, such as distilled models or one-step diffusion methods. It also mentions that while PipeFusion reduces communication and memory costs, it may face challenges in scenarios with extremely long sequences where communication overhead becomes less significant.

\item {\bf Theory assumptions and proofs}
    \item[] Question: For each theoretical result, does the paper provide the full set of assumptions and a complete (and correct) proof?
    \item[] Answer: \answerNA{} % Replace by \answerYes{}, \answerNo{}, or \answerNA{}.
    \item[] Justification: The paper focuses on proposing and evaluating the PipeFusion method through experimental results, and does not include formal theoretical results, theorems, or proofs.

    \item {\bf Experimental result reproducibility}
    \item[] Question: Does the paper fully disclose all the information needed to reproduce the main experimental results of the paper to the extent that it affects the main claims and/or conclusions of the paper (regardless of whether the code and data are provided or not)?
    \item[] Answer:  \answerYes{} % Replace by \answerYes{}, \answerNo{}, or \answerNA{}.
    \item[] Justification: The paper provides sufficient details to reproduce the main experimental results. It describes the experimental setup, including the hardware configuration (8×L40 PCIe GPUs), software stack (PyTorch, CUDA Runtime, diffusers), and the models used (Pixart, Stable-Diffusion 3, Flux.1). It also explains the methodology for implementing PipeFusion and other parallel approaches, and provides results for various image resolutions and configurations. While the code and data are not explicitly provided, the detailed descriptions allow for replication of the experiments.

\item {\bf Open access to data and code}
    \item[] Question: Does the paper provide open access to the data and code, with sufficient instructions to faithfully reproduce the main experimental results, as described in supplemental material?
    \item[] Answer: \answerYes{} % Replace by \answerYes{}, \answerNo{}, or \answerNA{}.
    \item[] Justification: The paper cites the original sources for the models and datasets used, such as Pixart, Stable-Diffusion 3, and Flux.1. It also mentions the use of publicly available datasets like COCO Captions 2014 for evaluation. While it does not explicitly state the license for each asset, it is common practice in the field to assume that publicly available datasets and models are used under their respective standard licenses (e.g., CC-BY for COCO Captions). The paper does not release any new assets, so there is no need to provide licenses for derived assets.

\item {\bf Experimental setting/details}
    \item[] Question: Does the paper specify all the training and test details (e.g., data splits, hyperparameters, how they were chosen, type of optimizer, etc.) necessary to understand the results?
    \item[] Answer: \answerYes{} % Replace by \answerYes{}, \answerNo{}, or \answerNA{}.
    \item[] Justification:The paper cites the original sources for the models and datasets used, such as Pixart, Stable-Diffusion 3, and Flux.1. It also mentions the use of publicly available datasets like COCO Captions 2014 for evaluation. While it does not explicitly state the license for each asset, it is common practice in the field to assume that publicly available datasets and models are used under their respective standard licenses (e.g., CC-BY for COCO Captions). The paper does not release any new assets, so there is no need to provide licenses for derived assets.

\item {\bf Experiment statistical significance}
    \item[] Question: Does the paper report error bars suitably and correctly defined or other appropriate information about the statistical significance of the experiments?
    \item[] Answer: \answerNo{} % Replace by \answerYes{}, \answerNo{}, or \answerNA{}.
    \item[] Justification: The paper does not report error bars, confidence intervals, or other statistical significance tests for the experiments. It is because the results is stable among tests. It presents the average latency and FID scores for multiple runs but does not provide information on the variability or statistical significance of these results. 

\item {\bf Experiments compute resources}
    \item[] Question: For each experiment, does the paper provide sufficient information on the computer resources (type of compute workers, memory, time of execution) needed to reproduce the experiments?
    \item[] Answer:  \answerYes{} % Replace by \answerYes{}, \answerNo{}, or \answerNA{}.
    \item[] Justification:  The paper provides detailed information on the computer resources used for the experiments. It specifies the hardware configuration as 8×L40 PCIe GPUs with 48GB memory each and mentions the software stack including PyTorch 2.4.1 and CUDA Runtime 12.1.105. Additionally, it reports the average latency for multiple runs, which indicates the time of execution for each experiment.
    
\item {\bf Code of ethics}
    \item[] Question: Does the research conducted in the paper conform, in every respect, with the NeurIPS Code of Ethics \url{https://neurips.cc/public/EthicsGuidelines}?
    \item[] Answer: \answerYes{} % Replace by \answerYes{}, \answerNo{}, or \answerNA{}.
    \item[] Justification: The research described in the paper does not involve human subjects, sensitive data, or any other elements that would require special ethical considerations according to the NeurIPS Code of Ethics. The paper focuses on improving the efficiency of diffusion transformer models for image generation, which does not pose risks related to privacy, discrimination, security, or other ethical concerns outlined in the guidelines.

\item {\bf Broader impacts}
    \item[] Question: Does the paper discuss both potential positive societal impacts and negative societal impacts of the work performed?
    \item[] Answer: \answerNo{} % Replace by \answerYes{}, \answerNo{}, or \answerNA{}.
    \item[] Justification: The paper does not discuss the broader societal impacts of the work performed. While it focuses on improving the efficiency of diffusion transformer models for image generation, it does not address potential positive or negative societal impacts. The research is foundational and does not directly tie to specific applications or deployments that could have significant societal consequences.
    
\item {\bf Safeguards}
    \item[] Question: Does the paper describe safeguards that have been put in place for responsible release of data or models that have a high risk for misuse (e.g., pretrained language models, image generators, or scraped datasets)?
    \item[] Answer: \answerNA{} % Replace by \answerYes{}, \answerNo{}, or \answerNA{}.
    \item[] Justification: The paper does not release any datasets or models that pose a high risk for misuse. It focuses on the methodology and experimental results of the PipeFusion approach without distributing any models or datasets that could be misused.

\item {\bf Licenses for existing assets}
    \item[] Question: Are the creators or original owners of assets (e.g., code, data, models), used in the paper, properly credited and are the license and terms of use explicitly mentioned and properly respected?
    \item[] Answer: \answerYes{} % Replace by \answerYes{}, \answerNo{}, or \answerNA{}.
    \item[] Justification: The paper cites the original sources for the models and datasets used, such as Pixart, Stable-Diffusion 3, and Flux.1. It also mentions the use of publicly available datasets like COCO Captions 2014 for evaluation. While it does not explicitly state the license for each asset, it is common practice in the field to assume that publicly available datasets and models are used under their respective standard licenses (e.g., CC-BY for COCO Captions). The paper does not release any new assets, so there is no need to provide licenses for derived assets.

\item {\bf New assets}
    \item[] Question: Are new assets introduced in the paper well documented and is the documentation provided alongside the assets?
    \item[] Answer:  \answerNA{} % Replace by \answerYes{}, \answerNo{}, or \answerNA{}.
    \item[] Justification: The paper does not introduce or release any new assets such as datasets, models, or code. Therefore, there is no need to document or provide documentation for new assets.

\item {\bf Crowdsourcing and research with human subjects}
    \item[] Question: For crowdsourcing experiments and research with human subjects, does the paper include the full text of instructions given to participants and screenshots, if applicable, as well as details about compensation (if any)? 
    \item[] Answer:  \answerNA{} % Replace by \answerYes{}, \answerNo{}, or \answerNA{}.
    \item[] Justification: The paper does not involve any crowdsourcing experiments or research with human subjects. It focuses on the technical aspects of parallelizing diffusion transformer models for image generation, which does not require human participation or data collection.

\item {\bf Institutional review board (IRB) approvals or equivalent for research with human subjects}
    \item[] Question: Does the paper describe potential risks incurred by study participants, whether such risks were disclosed to the subjects, and whether Institutional Review Board (IRB) approvals (or an equivalent approval/review based on the requirements of your country or institution) were obtained?
    \item[] Answer: \answerNA{} % Replace by \answerYes{}, \answerNo{}, or \answerNA{}.
    \item[] Justification: The paper does not involve any research with human subjects, so IRB approvals or equivalent reviews are not applicable.

\item {\bf Declaration of LLM usage}
    \item[] Question: Does the paper describe the usage of LLMs if it is an important, original, or non-standard component of the core methods in this research? Note that if the LLM is used only for writing, editing, or formatting purposes and does not impact the core methodology, scientific rigorousness, or originality of the research, declaration is not required.
    %this research? 
    \item[] Answer: \answerNo{}
    \item[] Justification: The paper does not describe the usage of Large Language Models (LLMs) as an important, original, or non-standard component of the core methods. The research focuses on the development and evaluation of the PipeFusion method for parallelizing diffusion transformer models, which is a technical contribution unrelated to LLMs.

\end{enumerate}

\appendix

% https://proceedings.neurips.cc/paper_files/paper/2024/file/9c3828adf1500f5de3c56f6550dfe43c-Paper-Conference.pdf

\section{Image Quality}
\label{sec:app_image_quality}

\begin{figure*}[htbp!]
  \centering
  \small
  \begin{minipage}{0.16\textwidth}
    \centering
  \begin{normalsize}Original\end{normalsize}\\
  1 Device\\
  % Latency: 3.340s\\
% Inception Score 39,45649337768555
  FID: 22.78
% SFID: 92.1799
% Precision: 0.6156
% Recall: 0. 4212
  \end{minipage}
  \begin{minipage}{0.16\textwidth}
  \centering
  \begin{normalsize}DistriFusion\end{normalsize}\\
  4 Devices\\
  % Latency: 1.662s\\
  FID: 30.69
  \end{minipage}
  \begin{minipage}{0.16\textwidth }
  \centering
  \begin{normalsize}DistriFusion\end{normalsize}\\
  8 Devices\\
  % Latency: 4.428s\\
  FID: 41.81
  \end{minipage}
  \begin{minipage}{0.16\textwidth}
  \centering
  \begin{normalsize}PipeFusion\end{normalsize}\\
  2 Devices\\
  patch number=2\\
  % Latency: 1.759s\\
  % FID: 
%   Inception Score: 34.249996185302734
  FID: 23.60
% sFID: 91.79367453574423
% Precision: 0.568
% Recall: 0.3678
  \end{minipage}
  \begin{minipage}{0.16\textwidth}
  \centering
  \begin{normalsize}PipeFusion\end{normalsize}\\
  4 Devices\\
  patch number=4\\
  % Latency: \textbf{1.541s}\\
  FID: 25.23
  \end{minipage}
  \begin{minipage}{0.16\textwidth}
  \centering
  \begin{normalsize}PipeFusion\end{normalsize}\\
  8 Devices\\
  patch number=8 \\
  % Latency: 1.989s\\
  FID: 28.46
  \end{minipage}
  \begin{subfigure}[b]{\textwidth}
    \centering
    \includegraphics[width=0.16\textwidth]{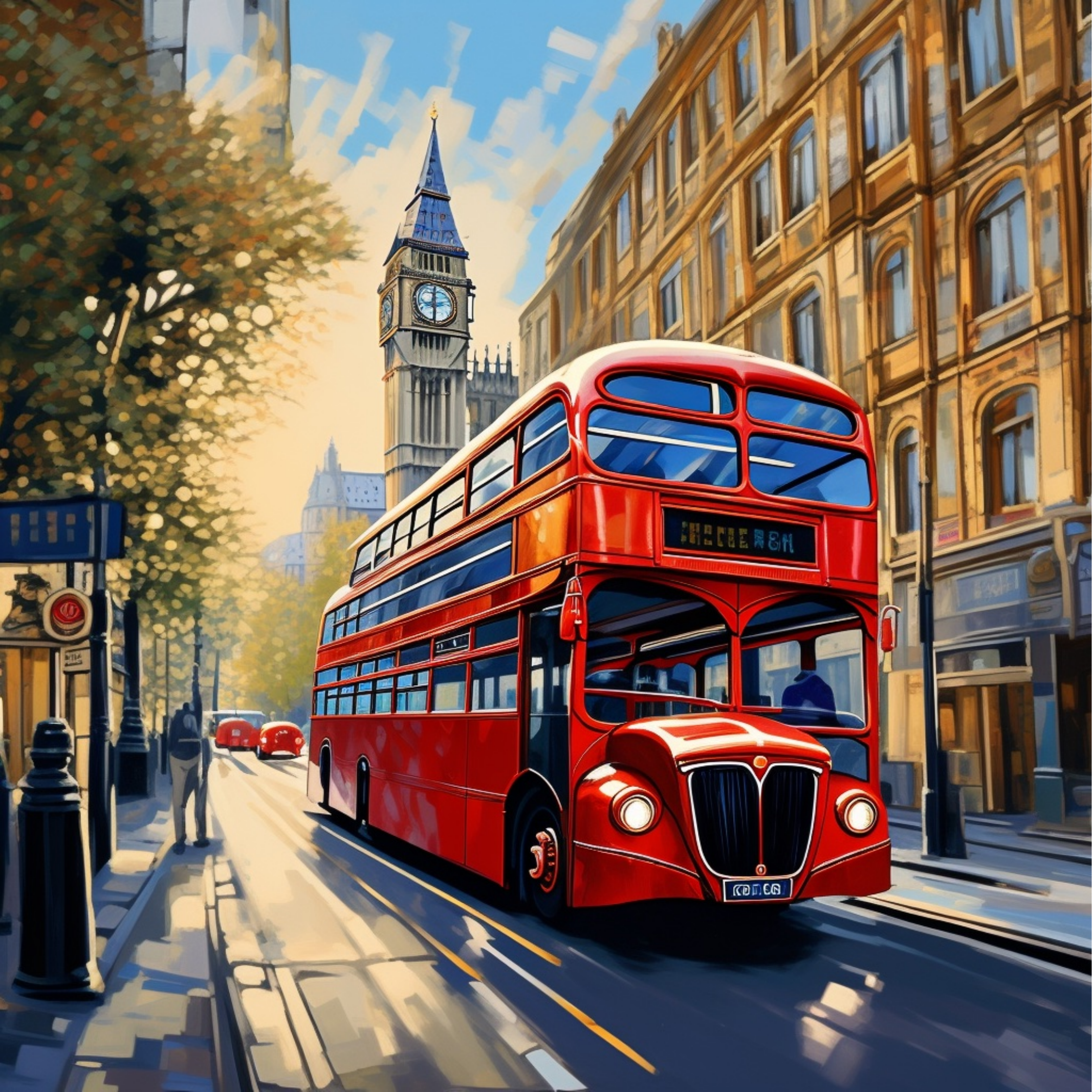}
    \includegraphics[width=0.16\textwidth]{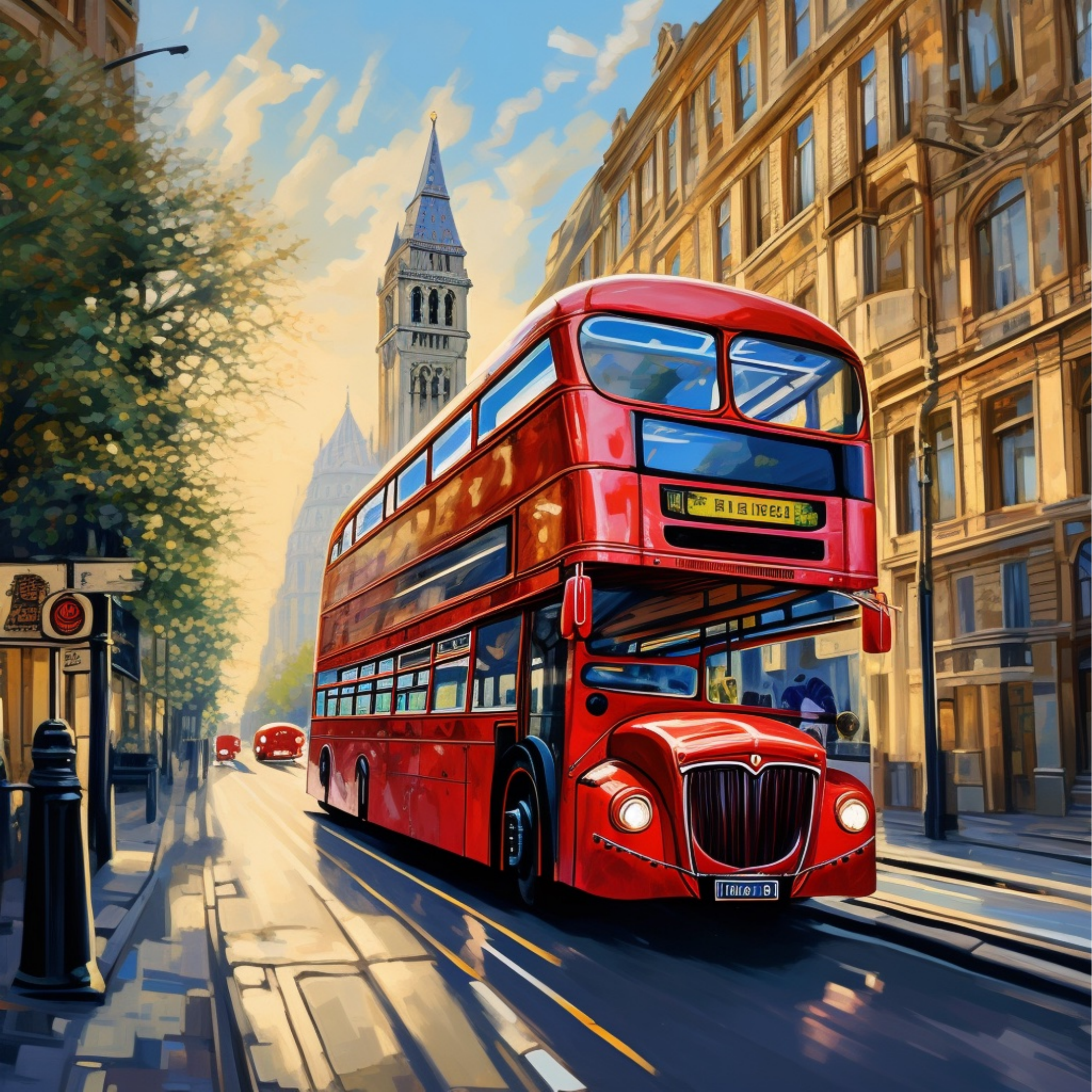}
    \includegraphics[width=0.16\textwidth]{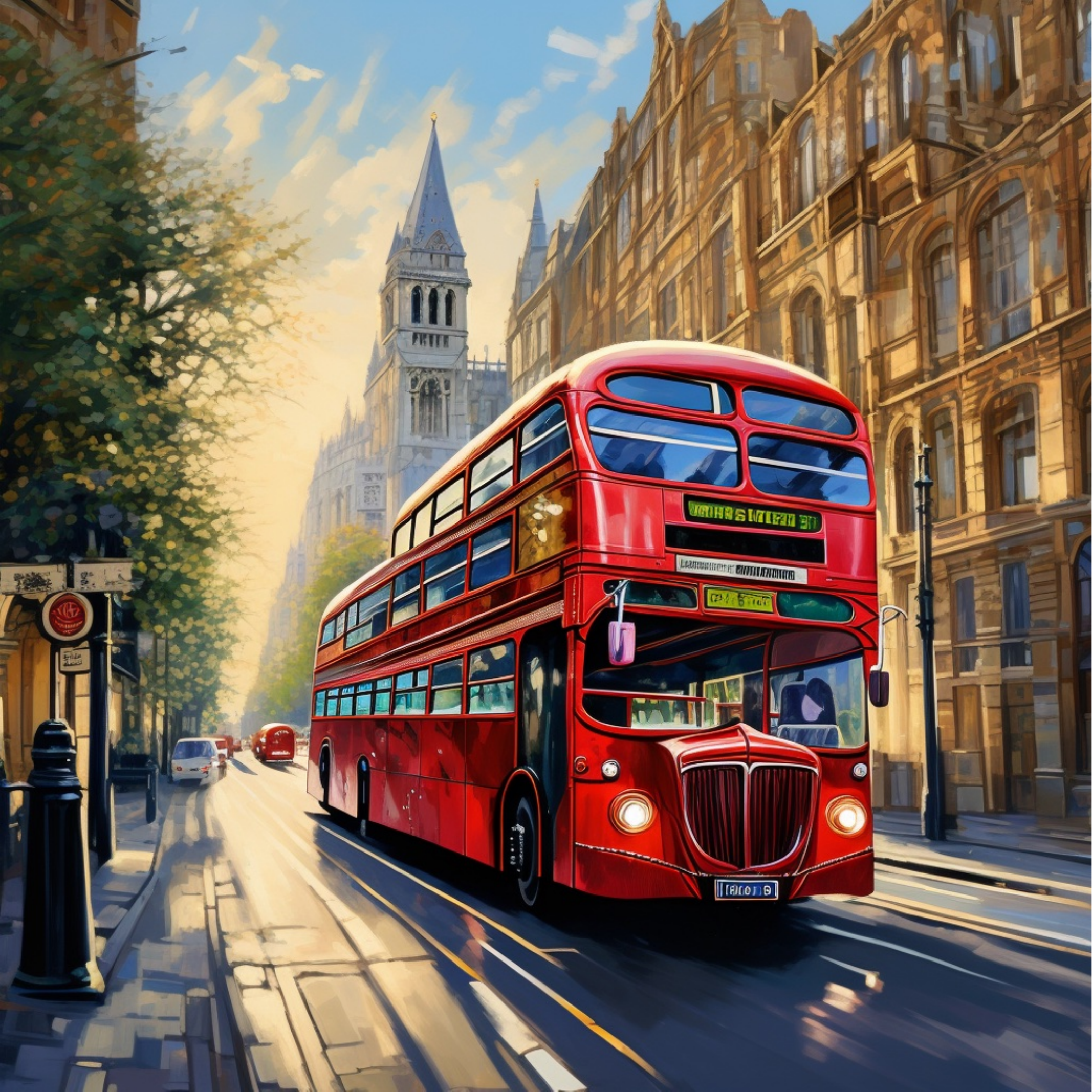}
    \includegraphics[width=0.16\textwidth]{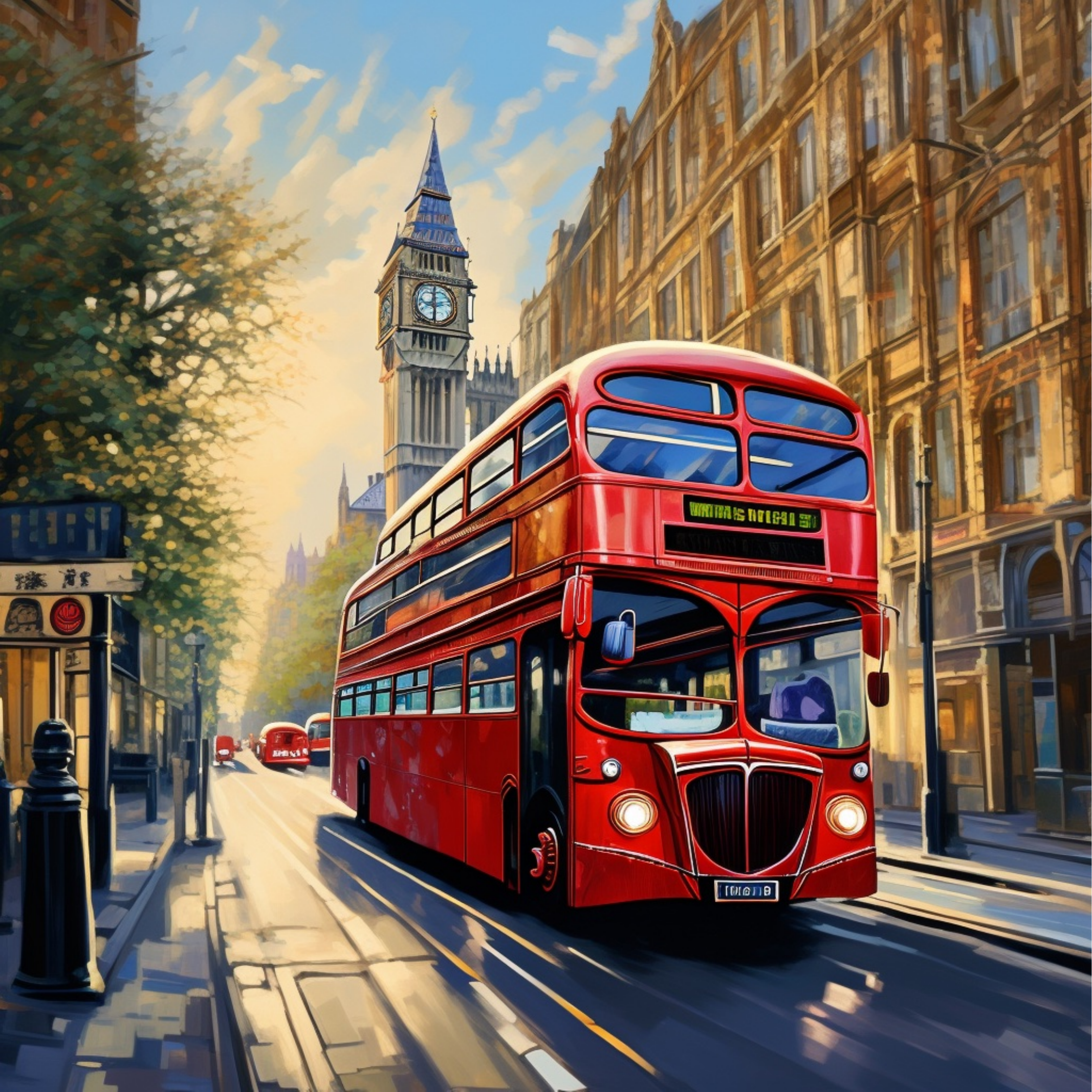}
    \includegraphics[width=0.16\textwidth]{./figures/results/bus/pip.pdf}
    \includegraphics[width=0.16\textwidth]{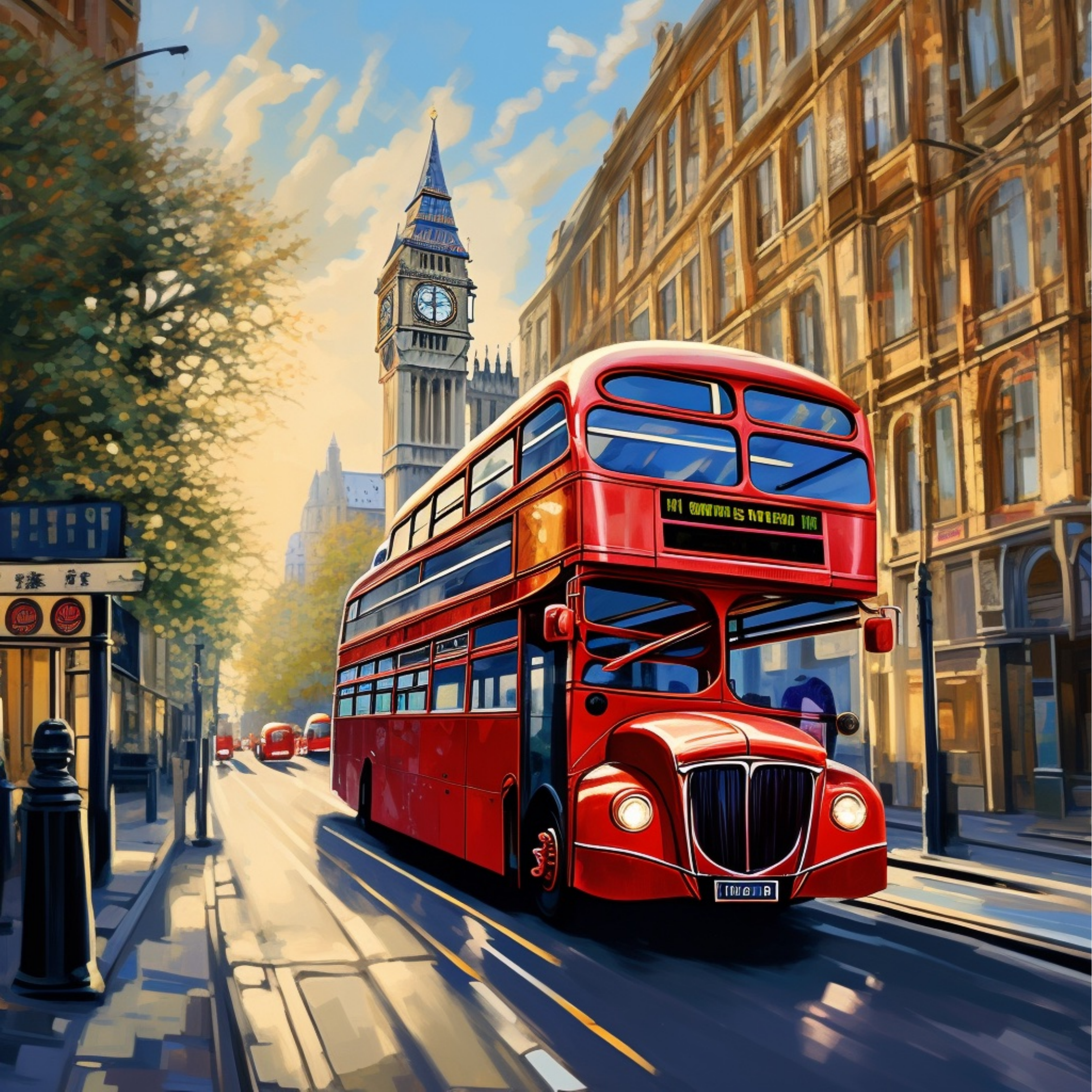}
    % \caption{Prompt: \textit{A double decker bus driving down the street.}}
    \parbox{0.9\textwidth}{\centering Prompt: \textit{A double decker bus driving down the street.}}
  \end{subfigure}
  \begin{subfigure}[b]{\textwidth}
    \centering
    \includegraphics[width=0.16\textwidth]{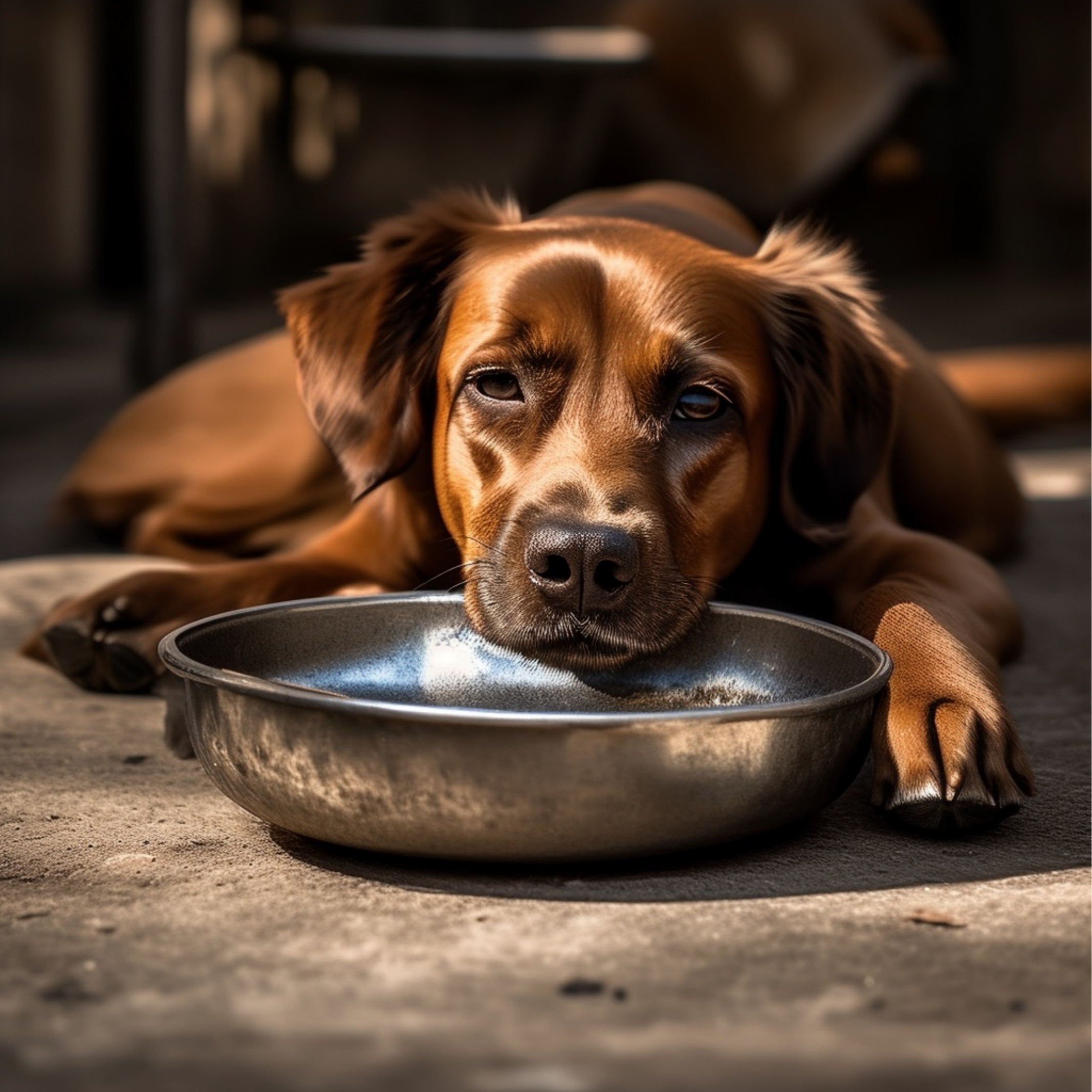}
    \includegraphics[width=0.16\textwidth]{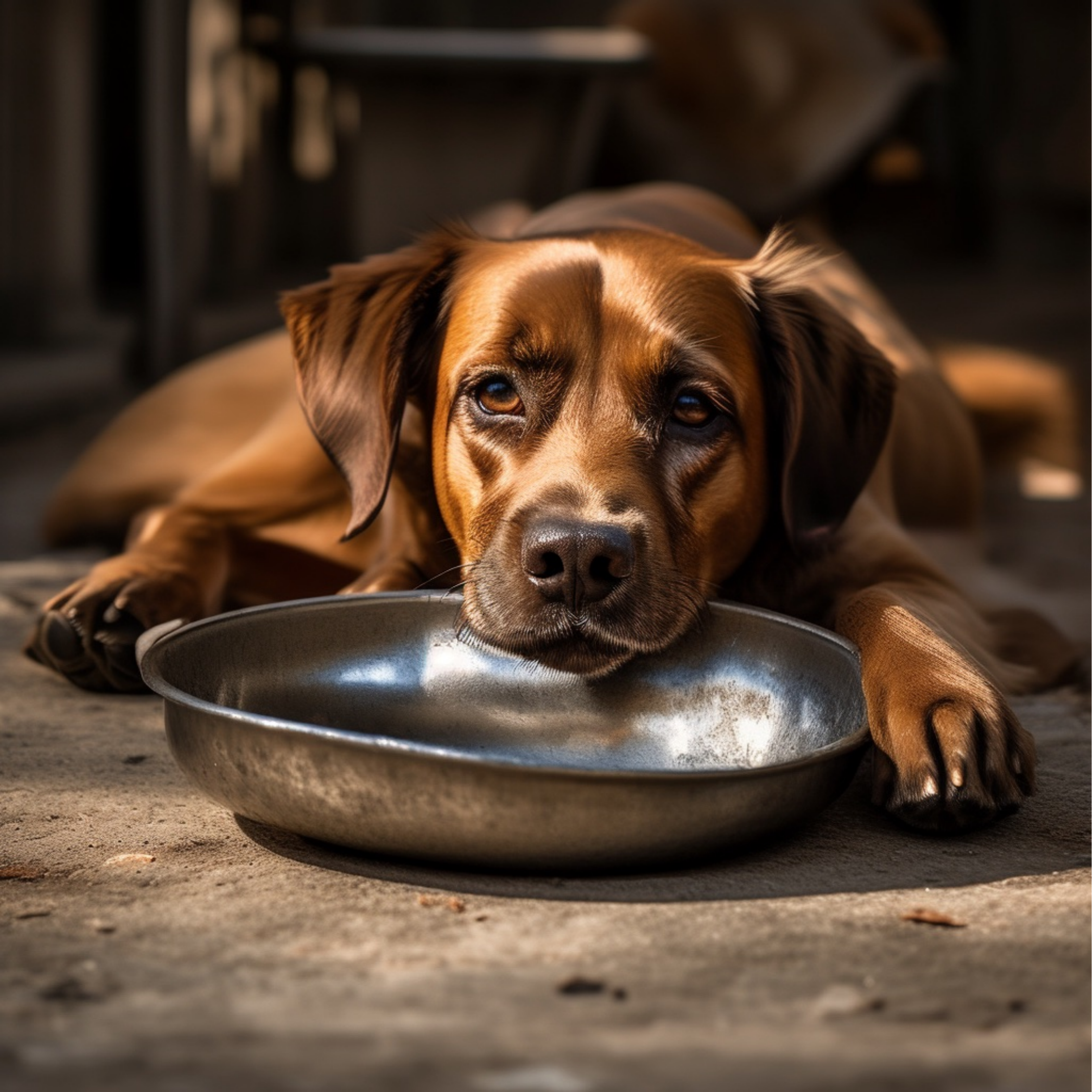}
    \includegraphics[width=0.16\textwidth]{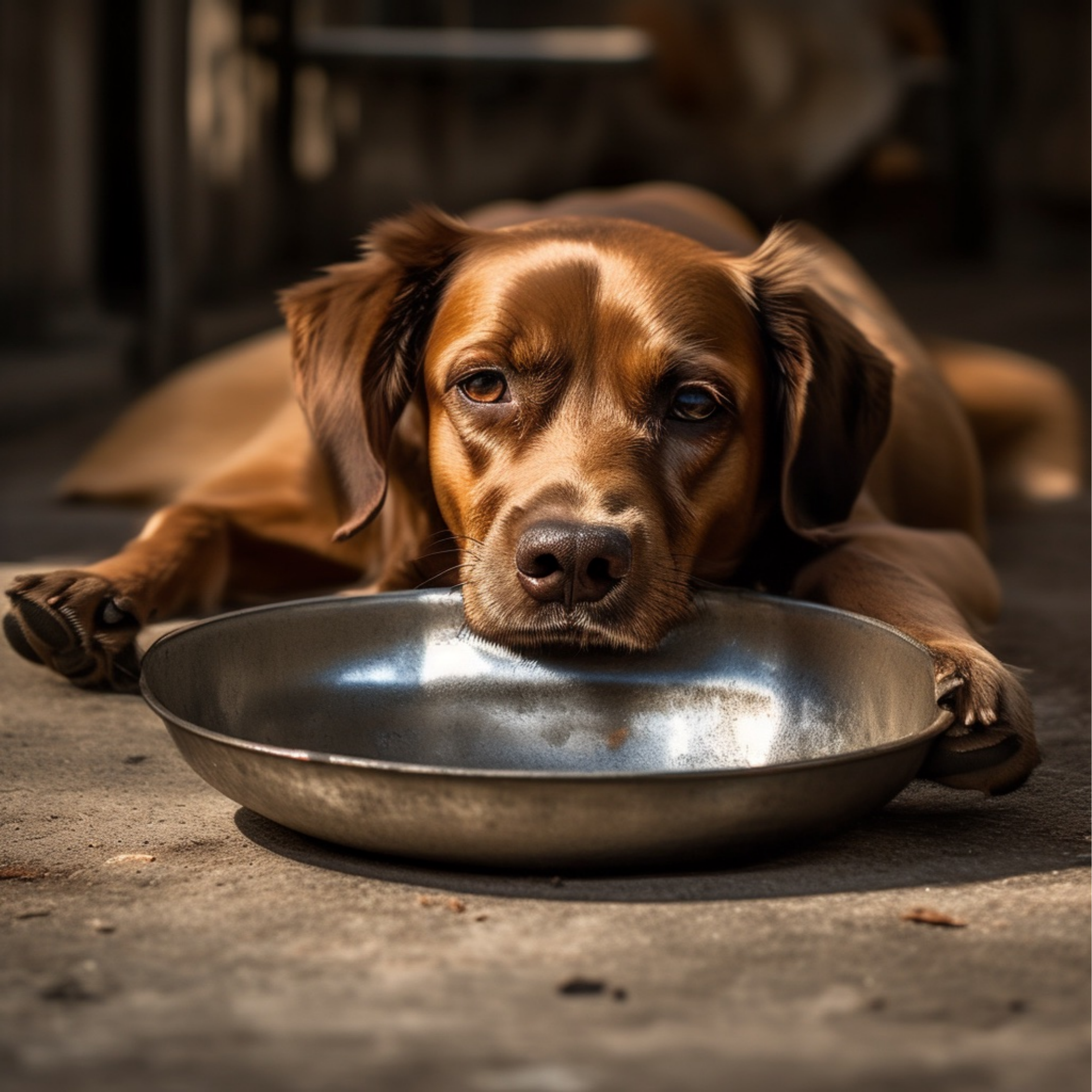}
    \includegraphics[width=0.16\textwidth]{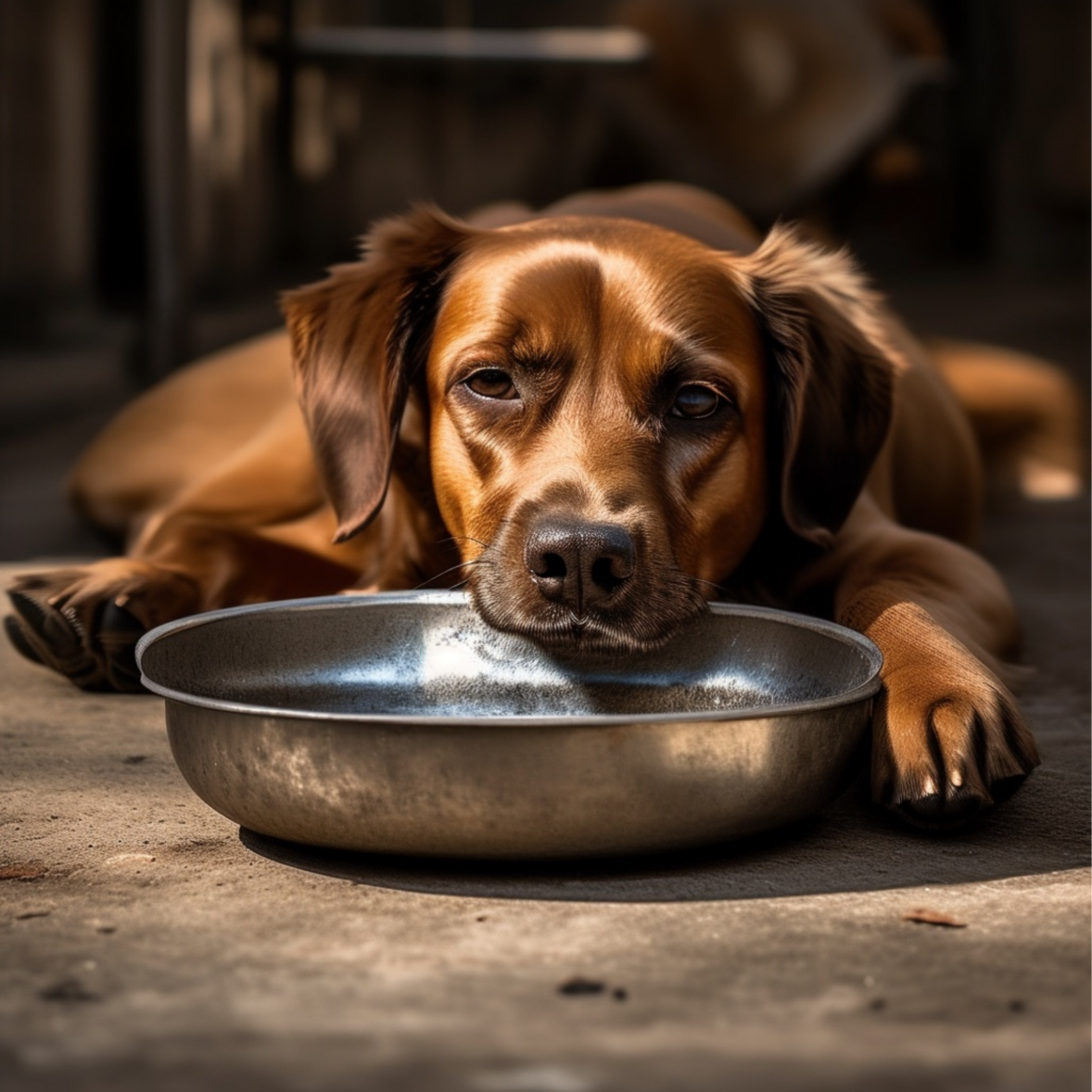}
    \includegraphics[width=0.16\textwidth]{./figures/results/dog/pip.pdf}
    \includegraphics[width=0.16\textwidth]{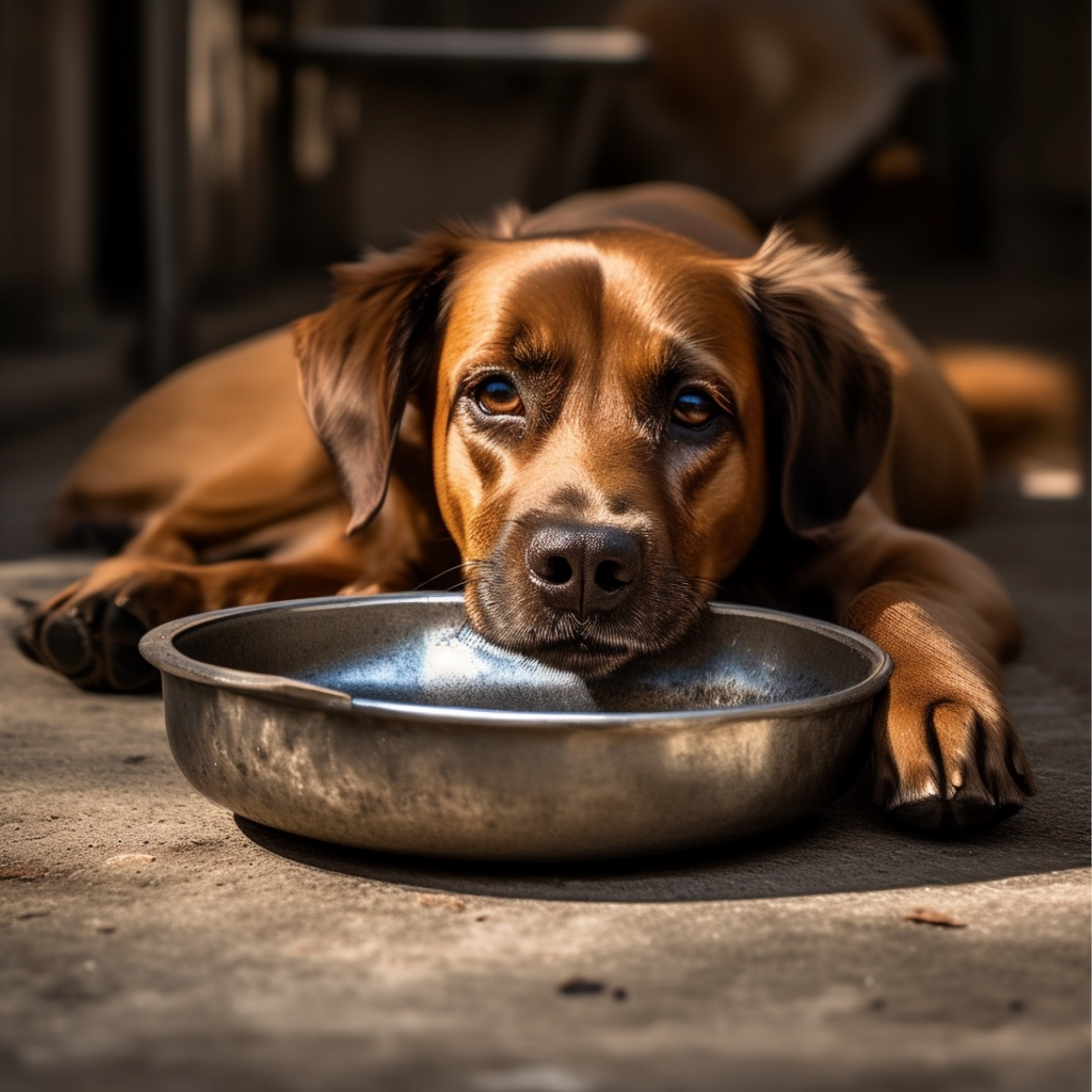}
    % \caption{Prompt: \textit{A brown dog laying on the ground with a metal bowl in front of him.}}
    \parbox{0.9\textwidth}{\centering Prompt: \textit{A brown dog laying on the ground with a metal bowl in front of him.}}
  \end{subfigure} 
  \begin{subfigure}[b]{\textwidth}
    \centering
    \includegraphics[width=0.16\textwidth]{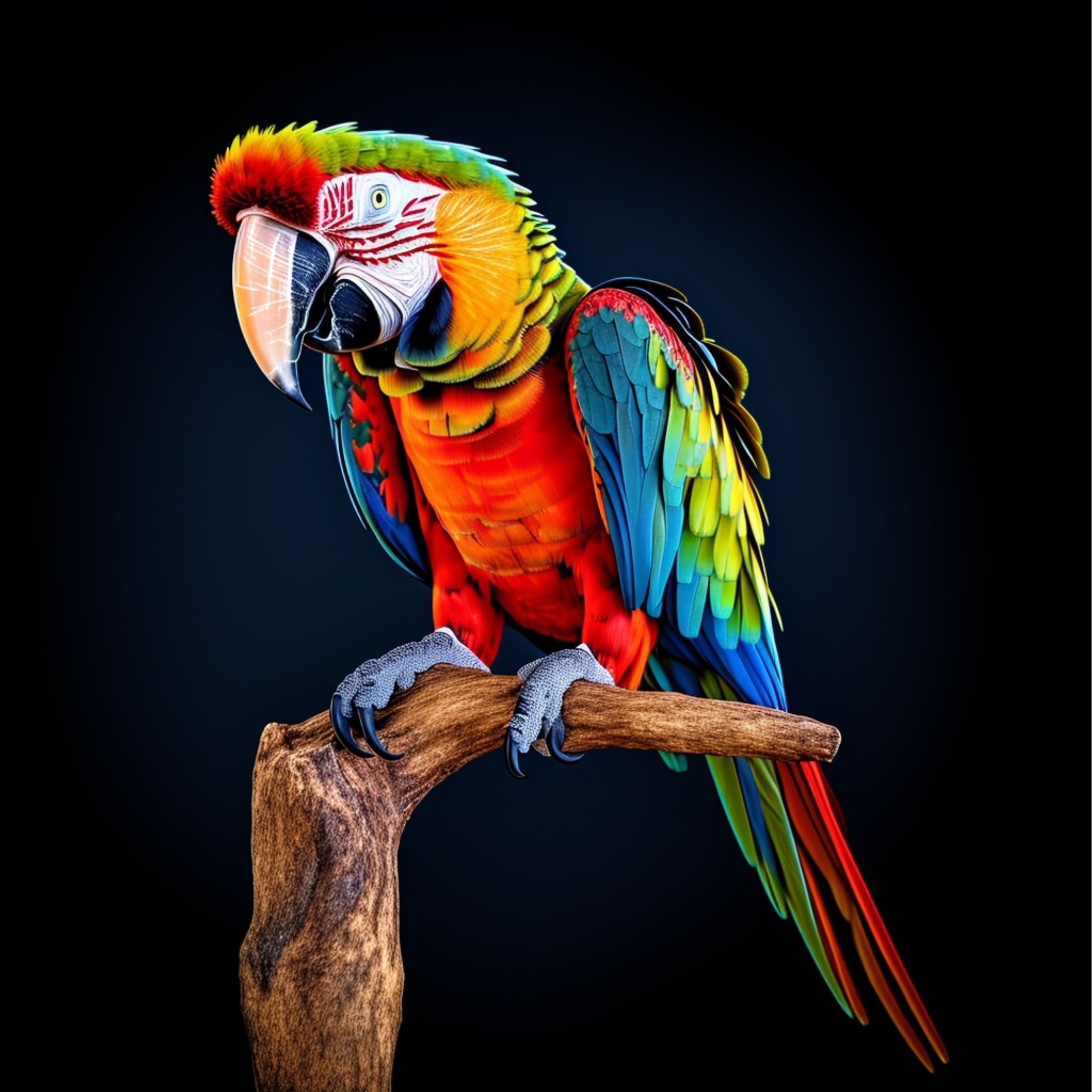}
    \includegraphics[width=0.16\textwidth]{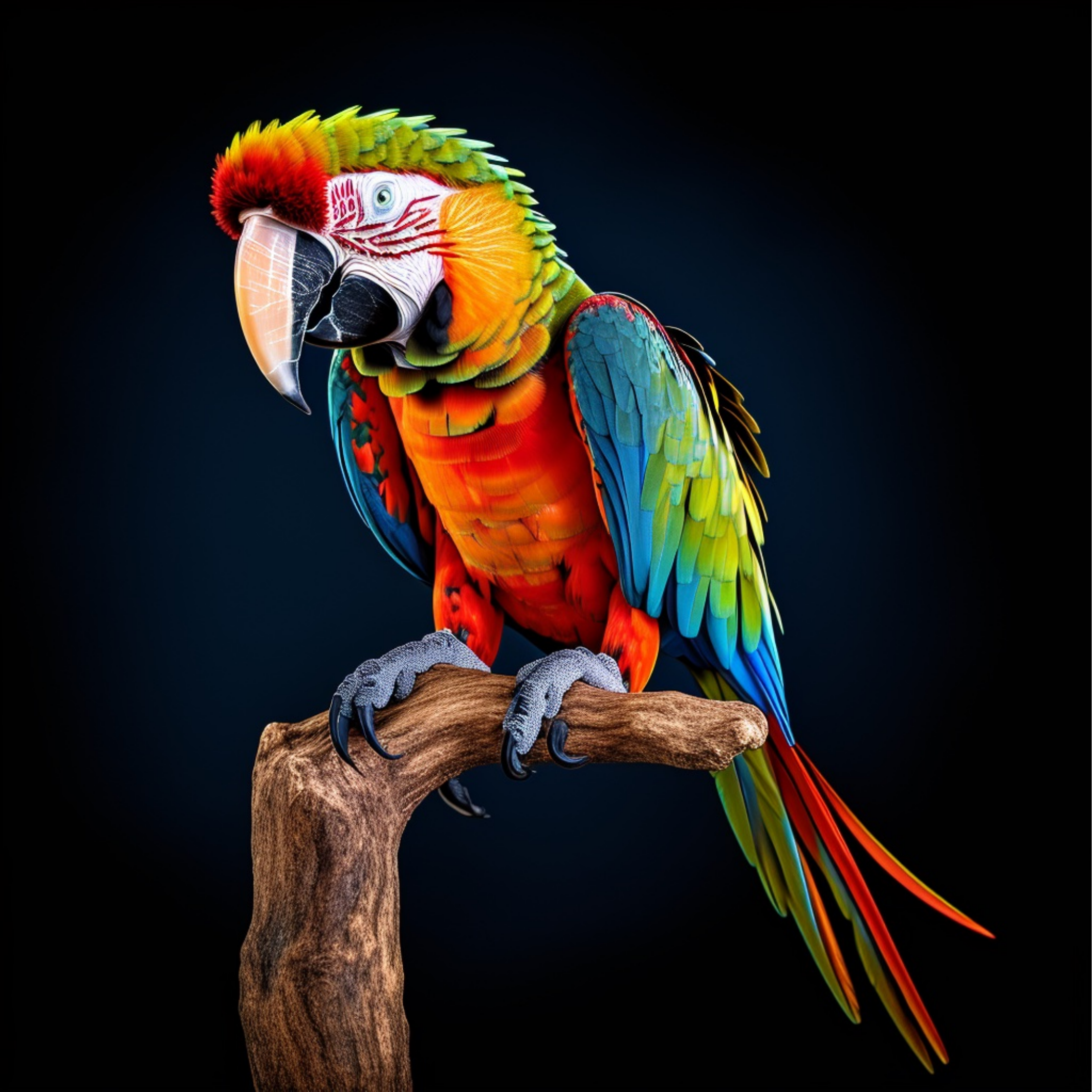}
    \includegraphics[width=0.16\textwidth]{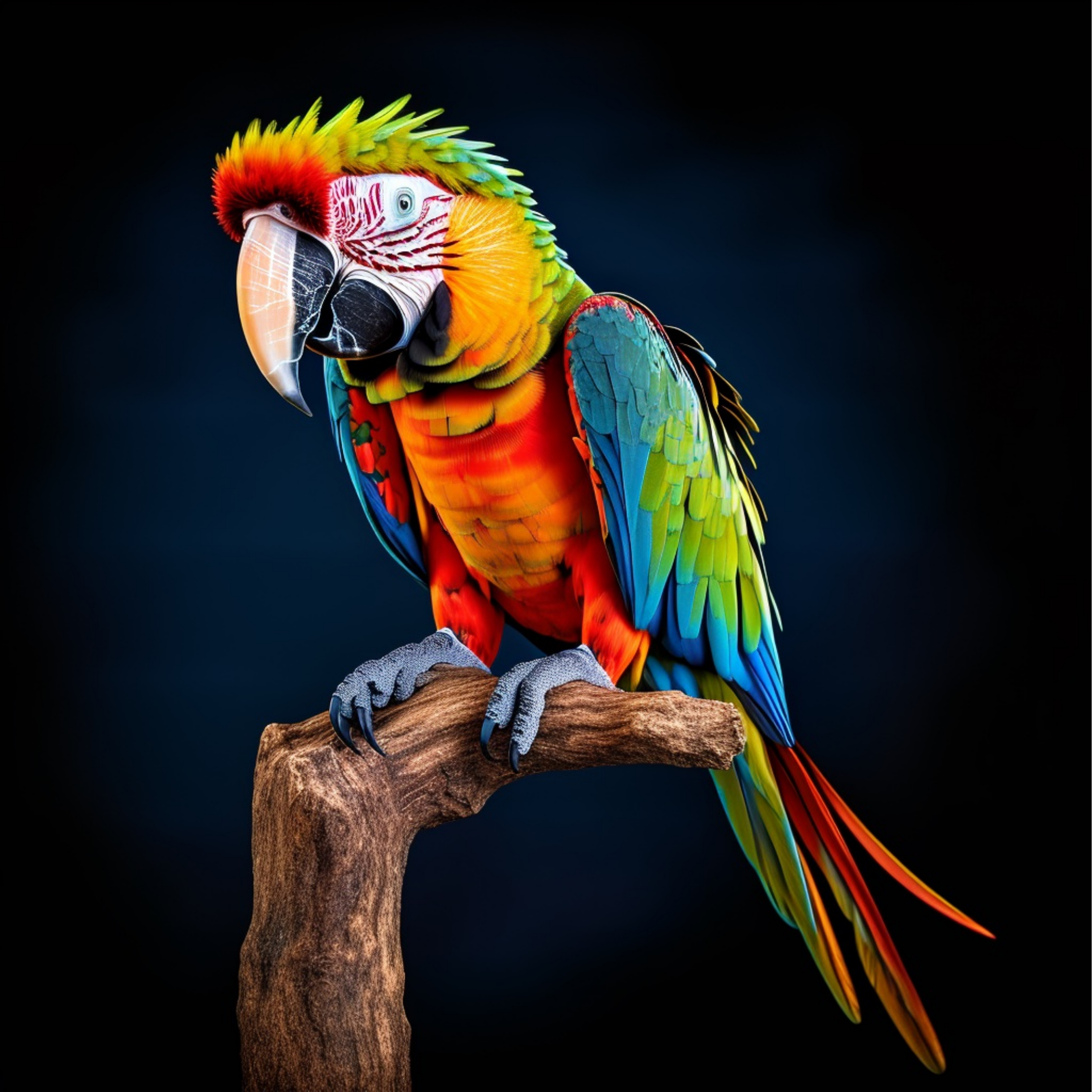}
    \includegraphics[width=0.16\textwidth]{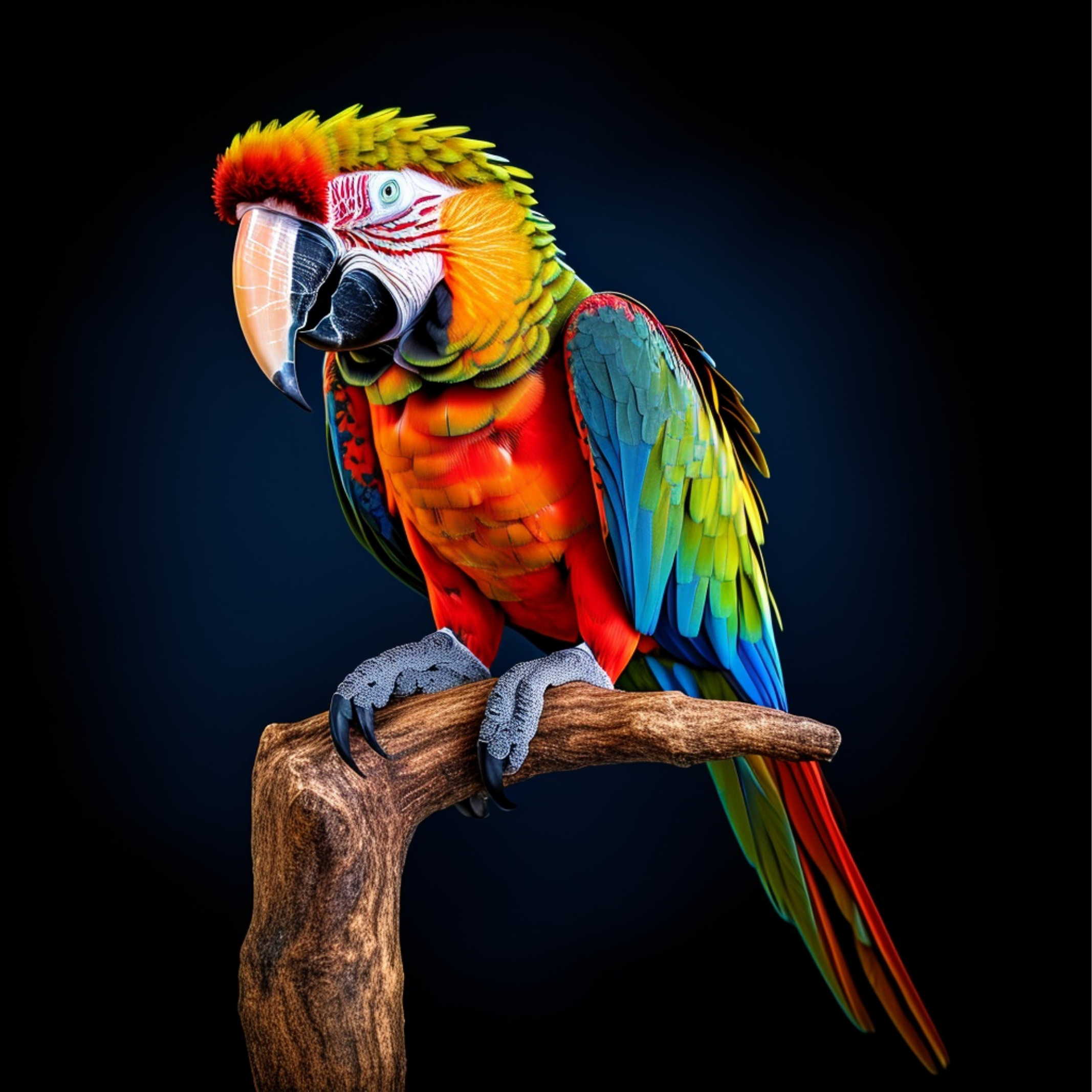}
    \includegraphics[width=0.16\textwidth]{./figures/results/parrot/pip.pdf}
    \includegraphics[width=0.16\textwidth]{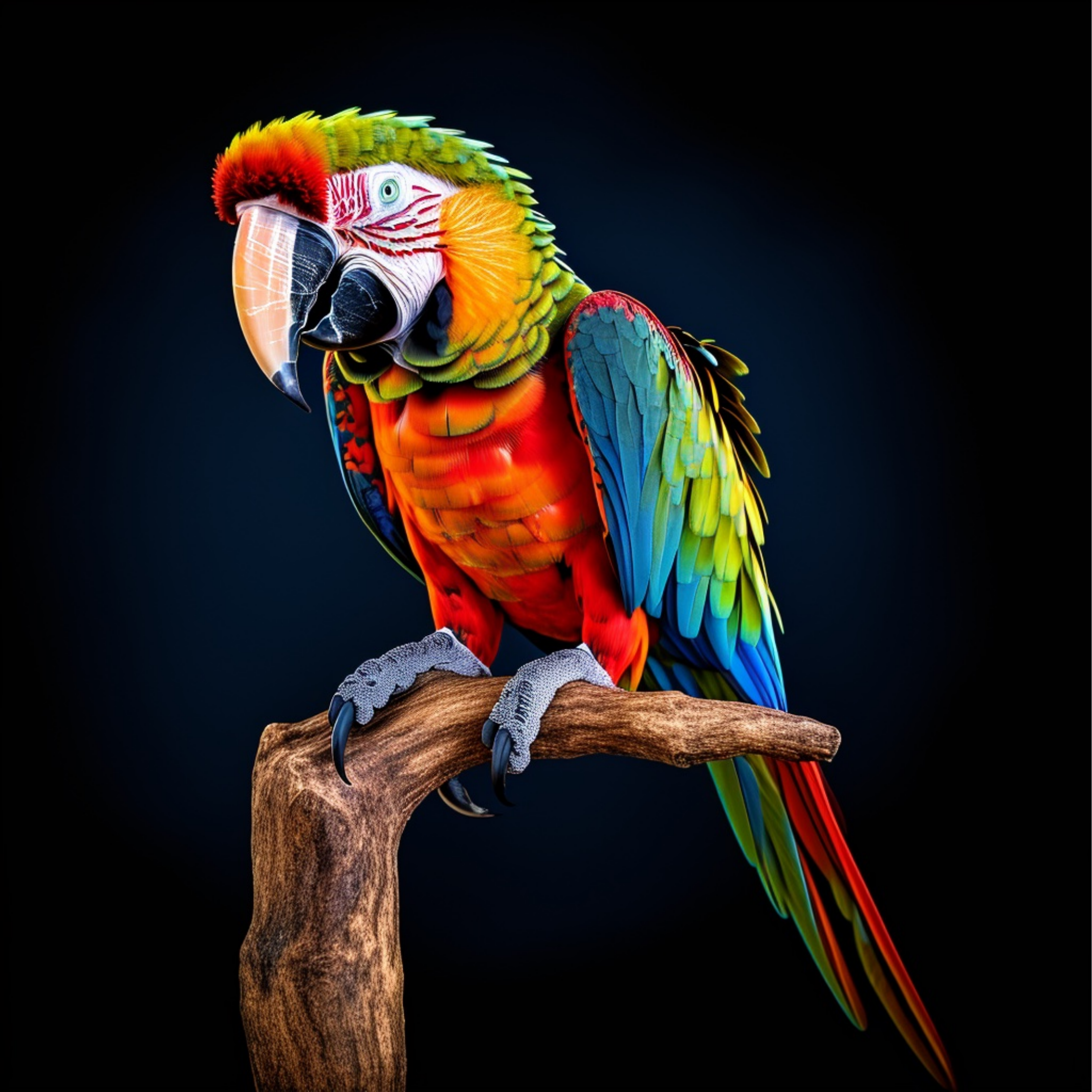}
    \parbox{0.9\textwidth}{\centering Prompt: \textit{A multi-colored parrot holding its foot up to its beak.}}
  \end{subfigure}
  \begin{subfigure}[b]{\textwidth}
    \centering
    \includegraphics[width=0.16\textwidth]{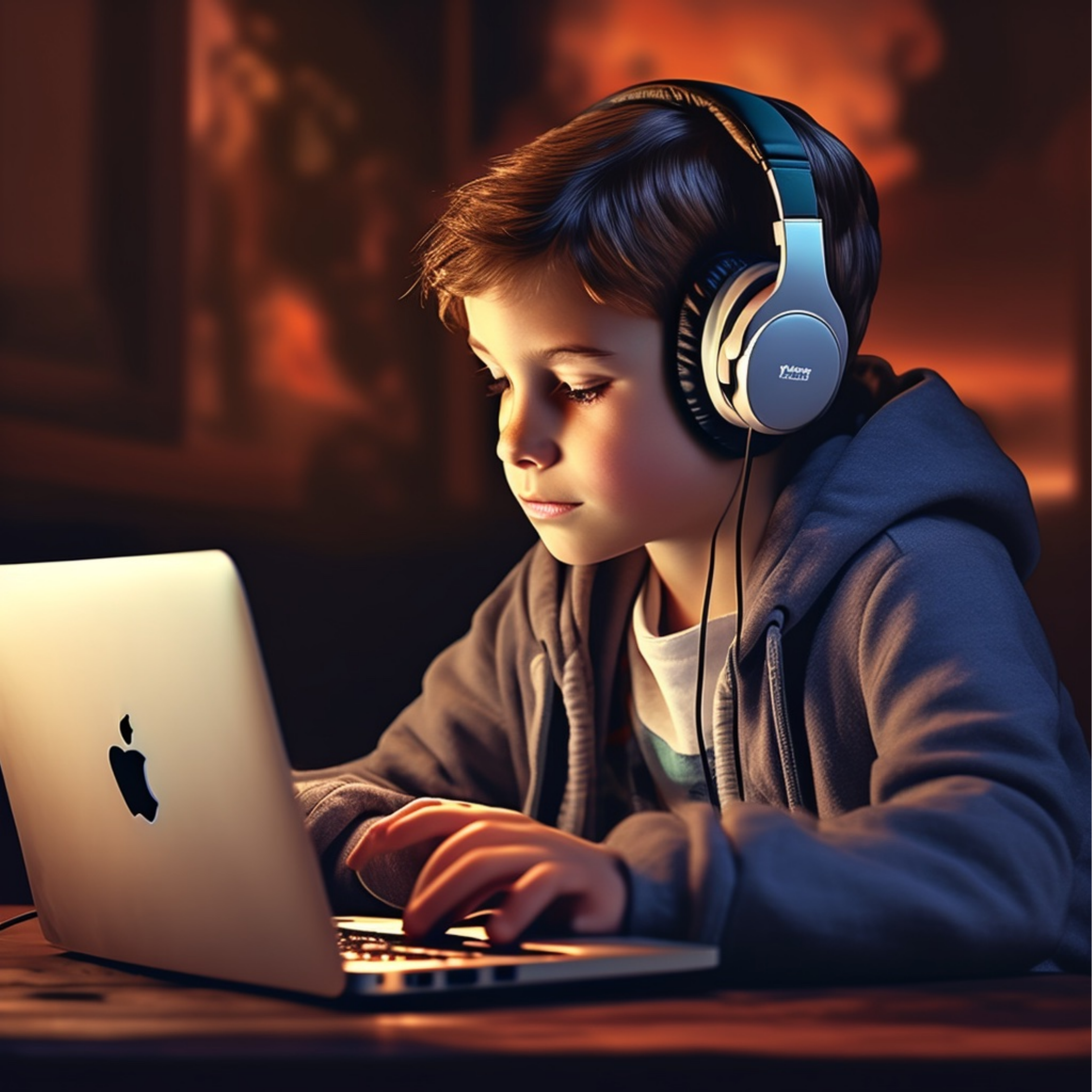}
    \includegraphics[width=0.16\textwidth]{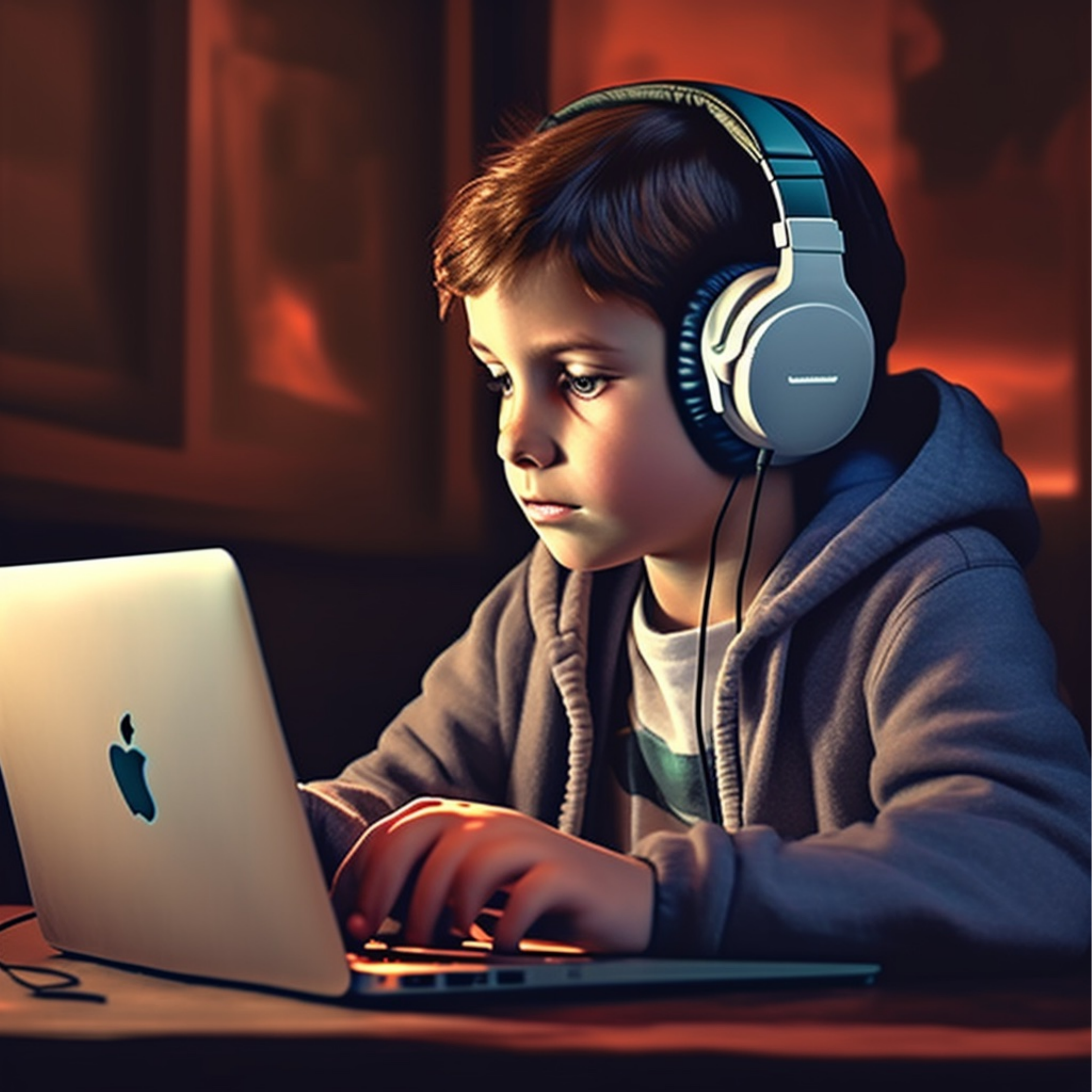}
    \includegraphics[width=0.16\textwidth]{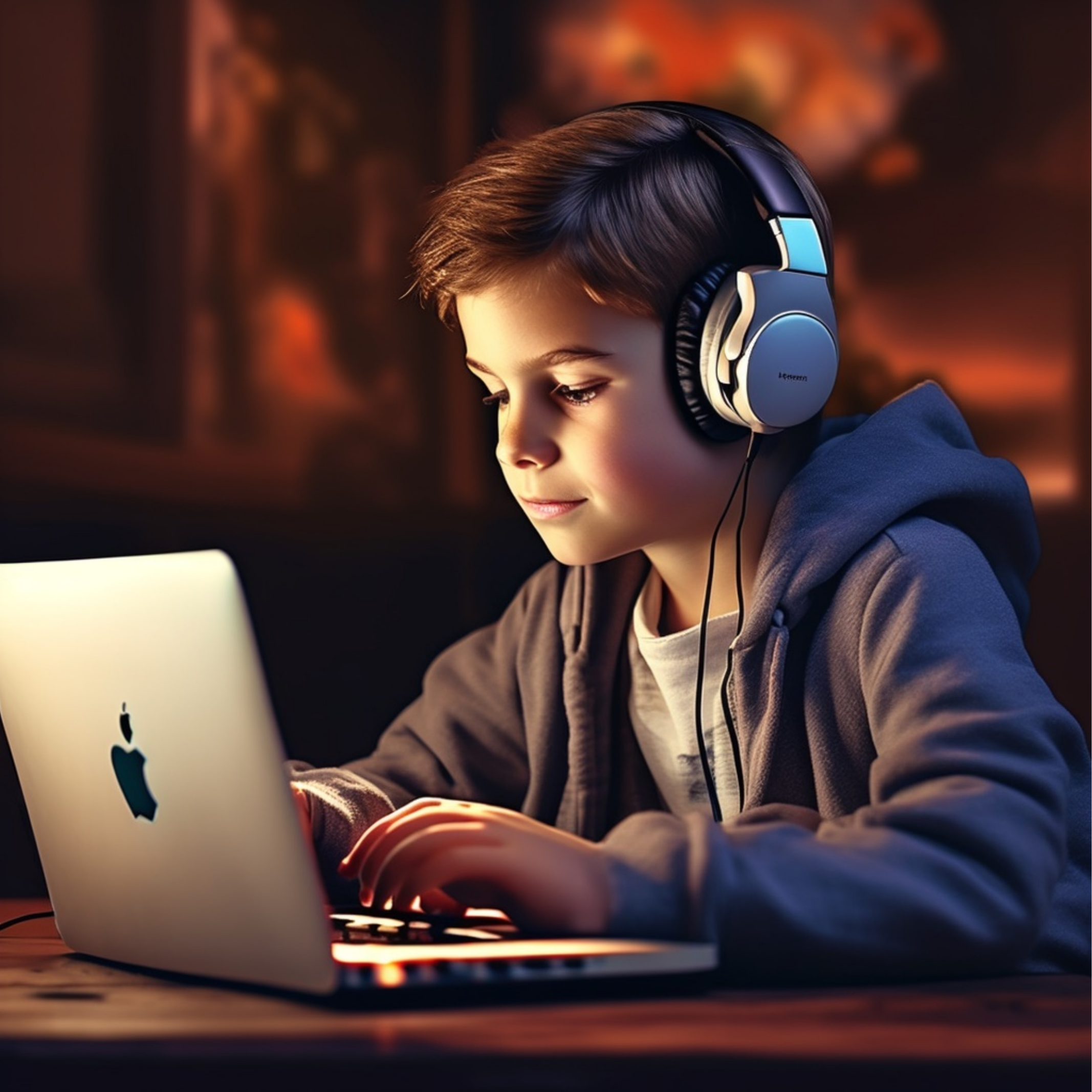}
    \includegraphics[width=0.16\textwidth]{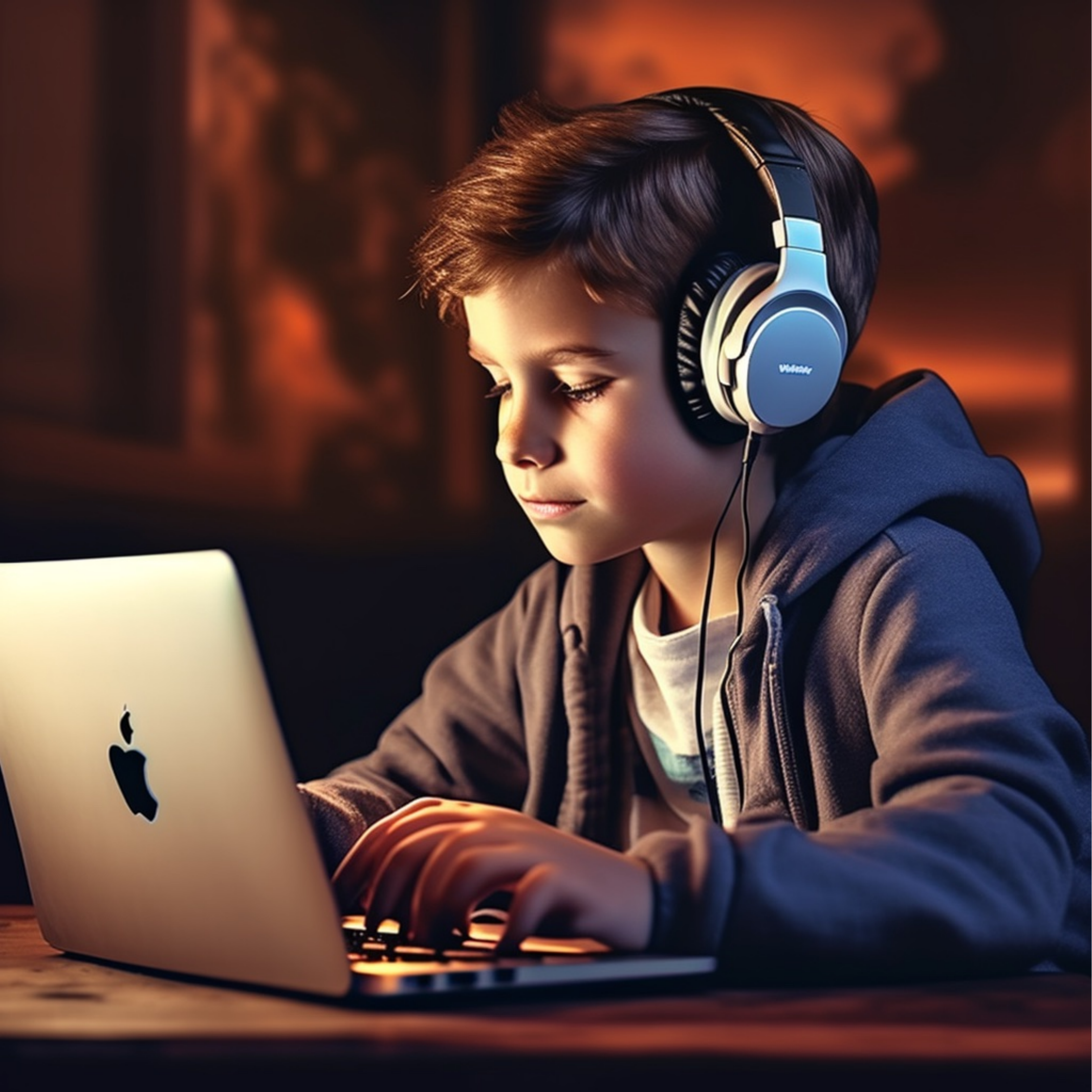}
    \includegraphics[width=0.16\textwidth]{./figures/results/kid/pip.pdf}
    \includegraphics[width=0.16\textwidth]{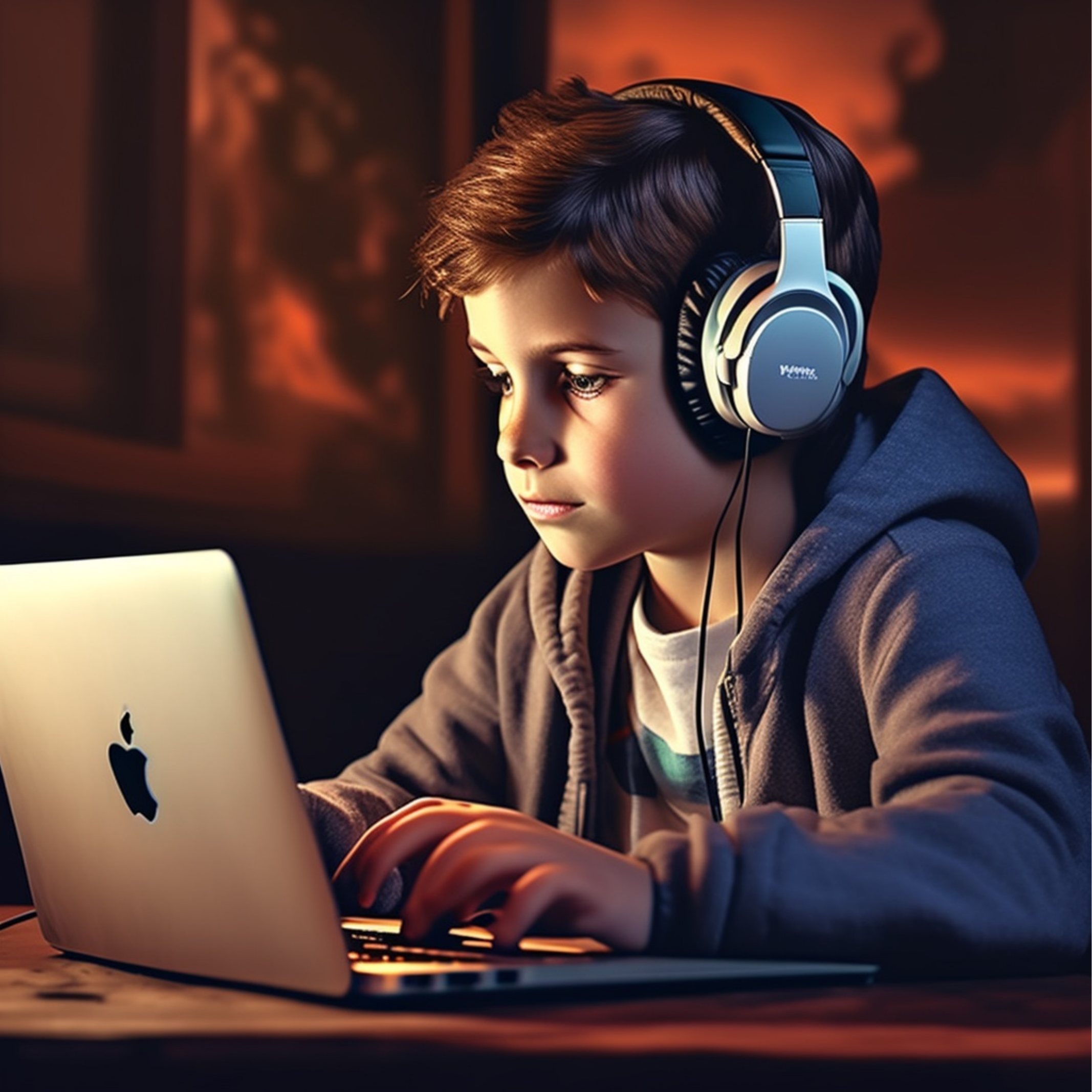}
    \parbox{0.9\textwidth}{\centering Prompt: \textit{A kid wearing headphones and using a laptop.}}
  \end{subfigure}
  \begin{subfigure}[b]{\textwidth}
    \centering
    \includegraphics[width=0.16\textwidth]{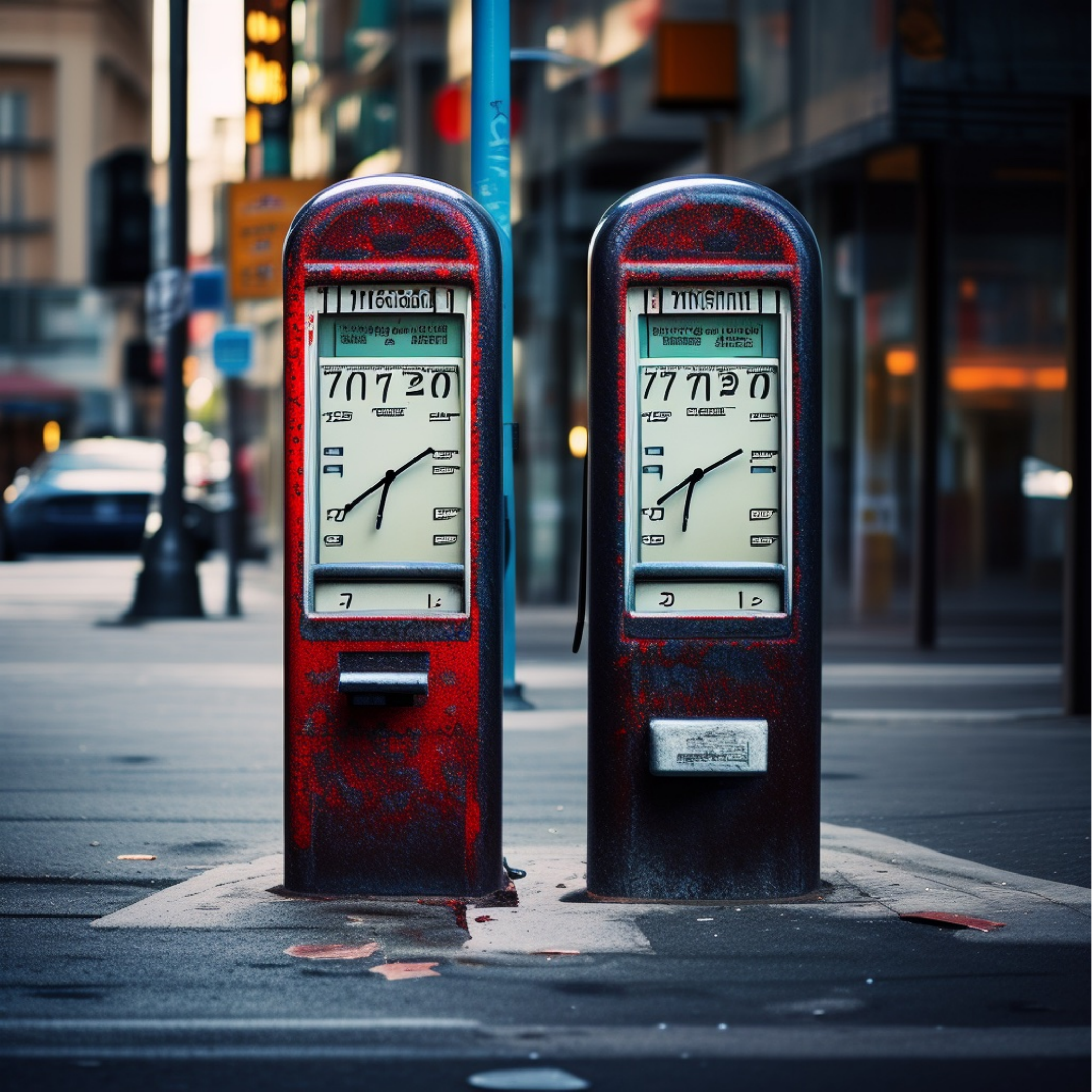}
    \includegraphics[width=0.16\textwidth]{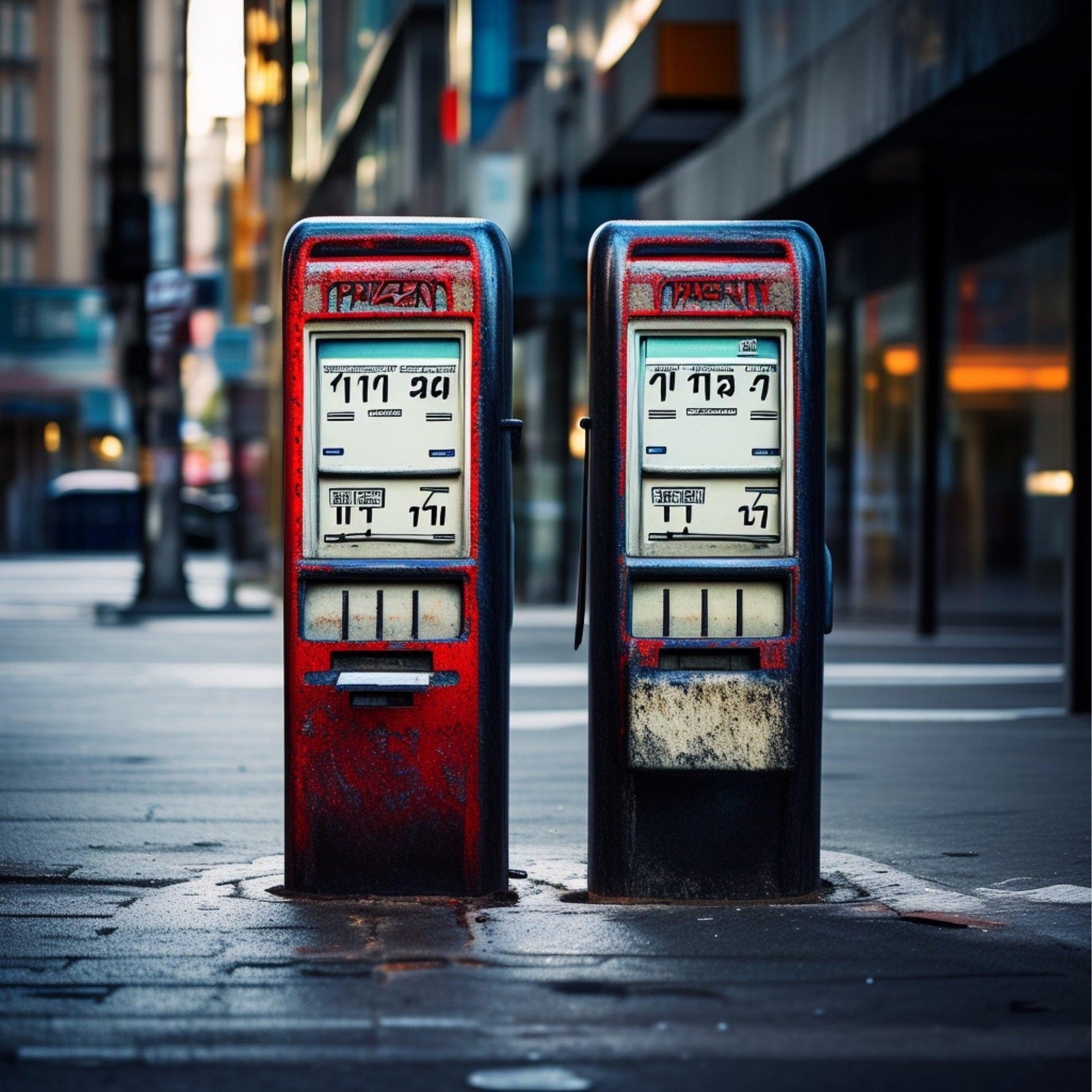}
    \includegraphics[width=0.16\textwidth]{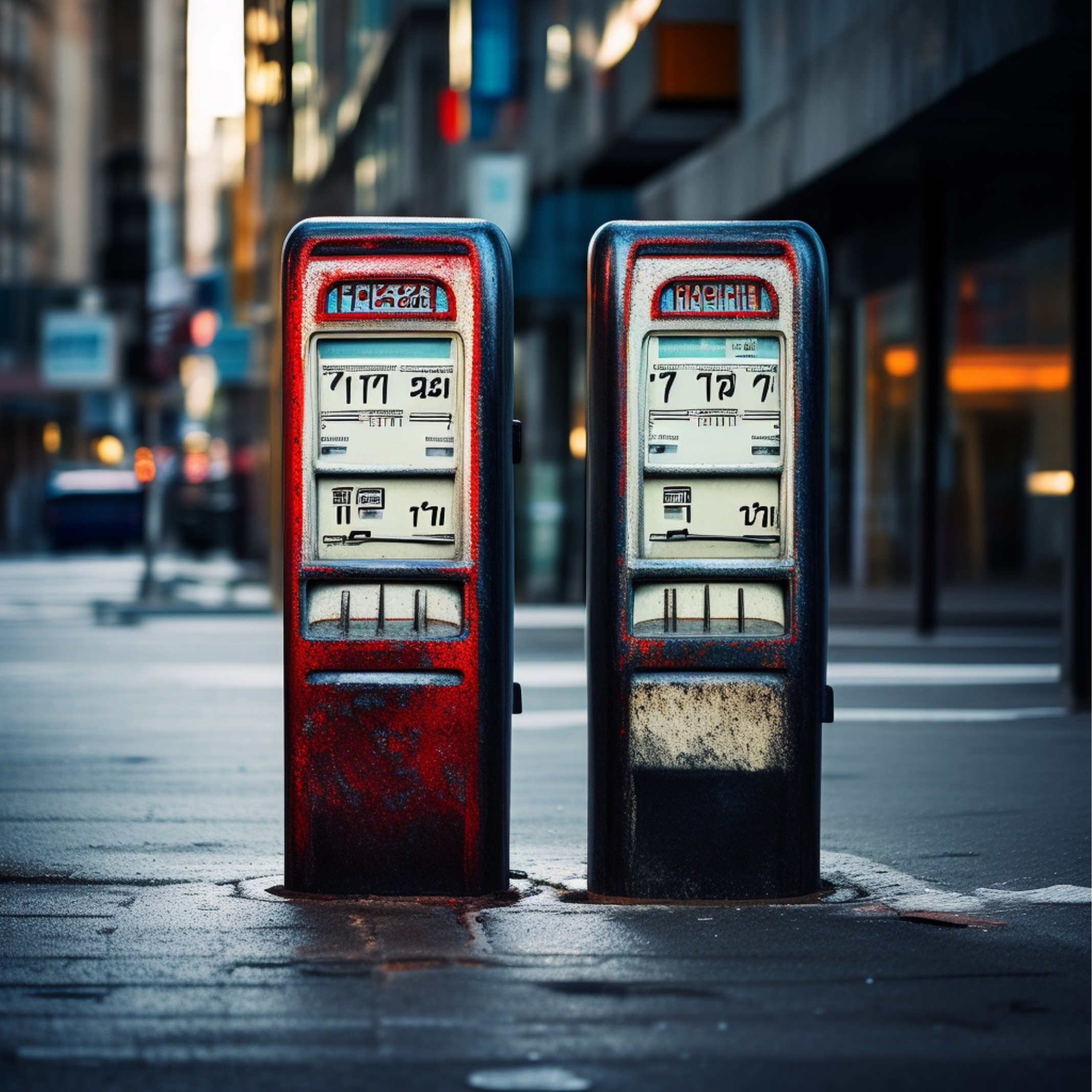}
    \includegraphics[width=0.16\textwidth]{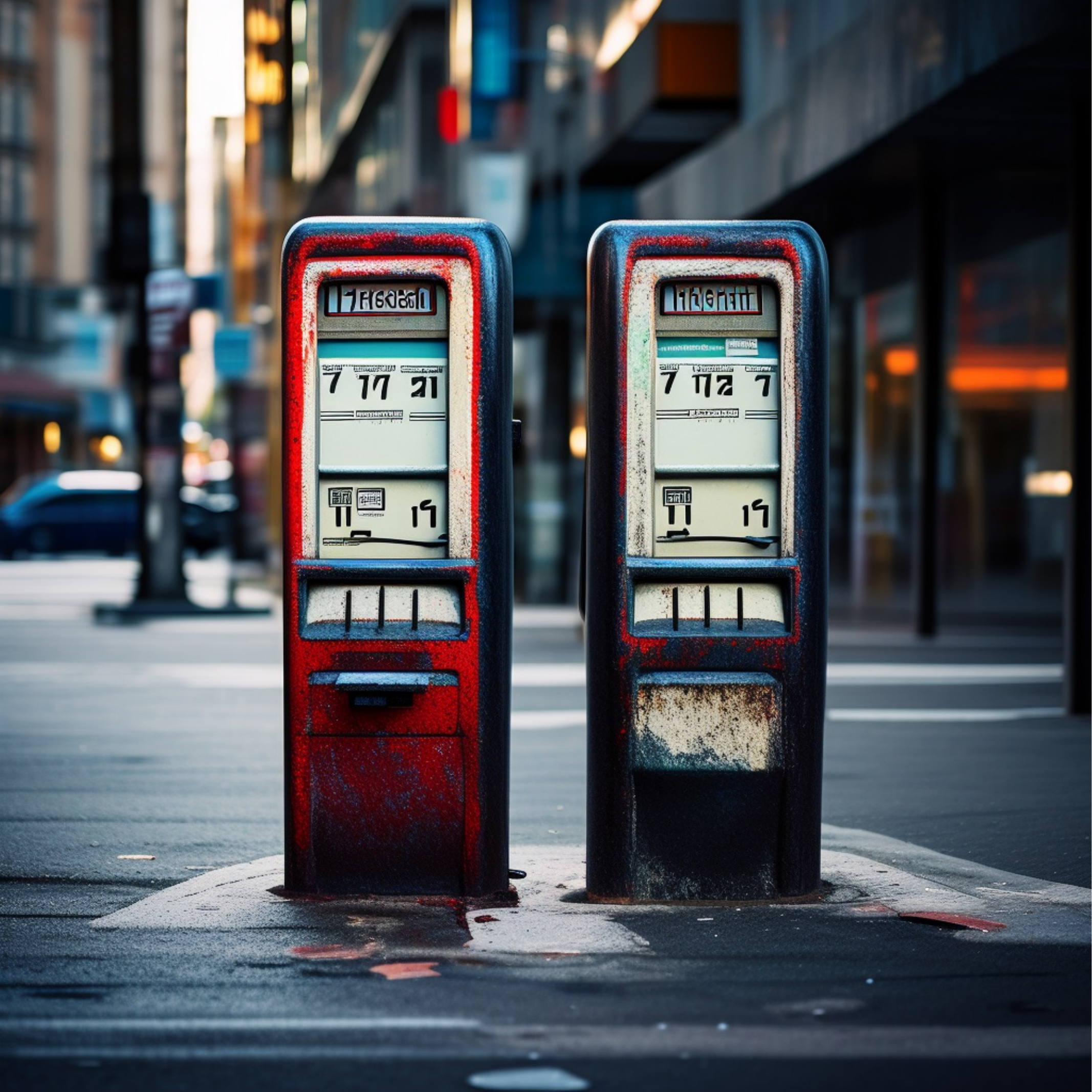}
    \includegraphics[width=0.16\textwidth]{./figures/results/parking/pip.pdf}
    \includegraphics[width=0.16\textwidth]{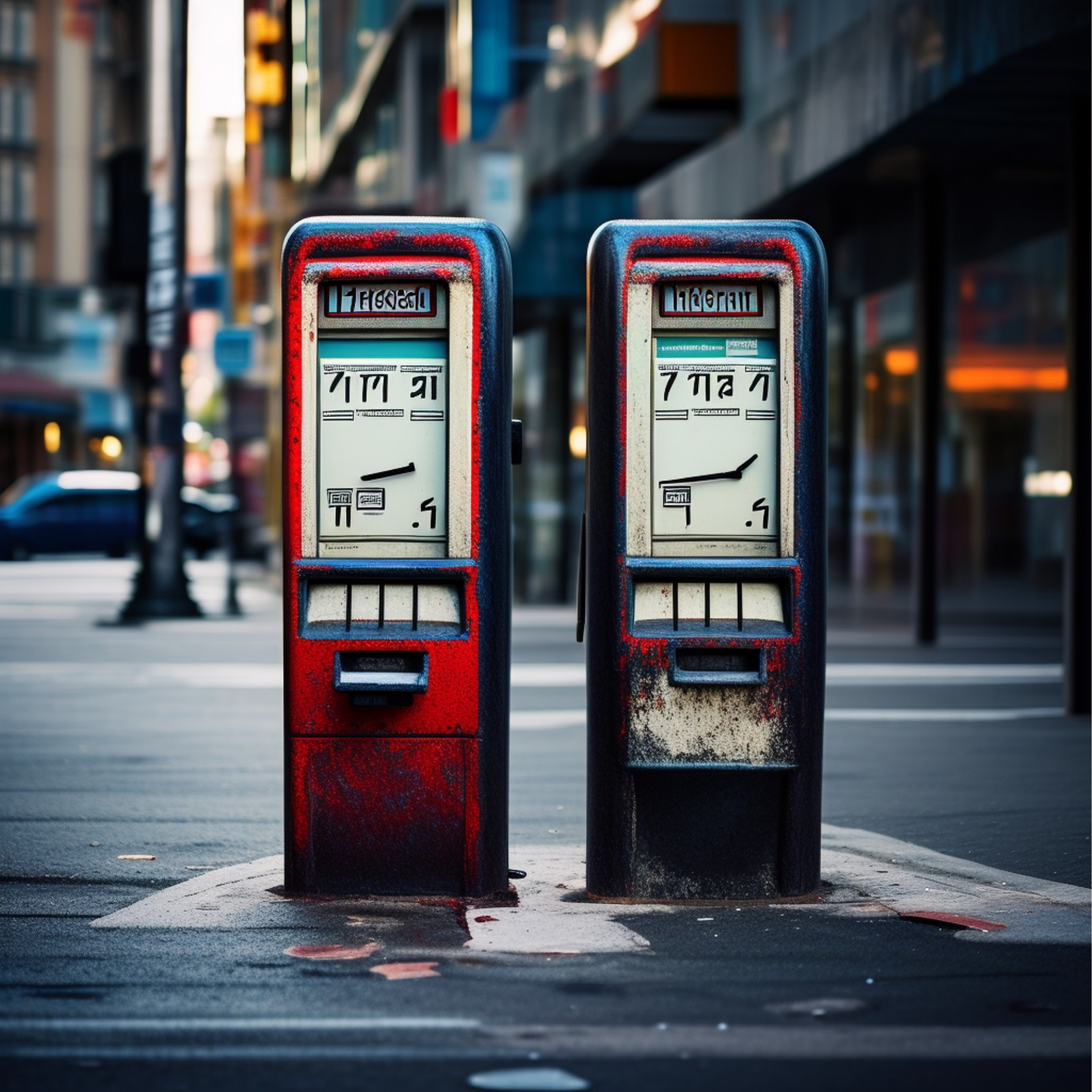}
    \parbox{0.9\textwidth}{\centering Prompt: \textit{A pair of parking meters reflecting expired times.}}
  \end{subfigure}
  % \begin{subfigure}[b]{\textwidth}
  %   \small
  %   \centering
  %   \begin{tabular}{cccccccccccc}
  %   \toprule
  %   \multicolumn{2}{c}{\textbf{Model}} & \multicolumn{6}{c}{\textbf{PixArt-XL-2-256-MS}} & \multicolumn{4}{c}{\textbf{FLUX.1[dev]}} \\ 
  %   \midrule
  %   \multicolumn{2}{c}{\textbf{Method}} & Original & \multicolumn{2}{c}{DistriFusion} & \multicolumn{3}{c}{PipeFusion} & Original & \multicolumn{3}{c}{PipeFusion} \\ % \hline
  %   \cmidrule(lr){3-3}\cmidrule(lr){4-5}\cmidrule(lr){6-8}\cmidrule(lr){9-9}\cmidrule{10-12} \multicolumn{2}{c}{\textbf{Device Number}} & 1 & 4 & 8 & 2 & 4 & 8 & 1 & 2 & 4 & 8 \\ % \hline
  %   \midrule
  %   \multirow{2}{*}{\textbf{FID $\downarrow$}}
  %   & \textbf{w/ G.T.} & 22.78 & 30.69 & 41.81 & 23.60 & 25.23 & 28.46 & 23.52 & 23.30 & 24.17 & 25.97 \\ % \hline
  %   & \textbf{w/ Orig.} & - & 8.41 & 19.81 & 1.67 & 3.11 & 6.09 & - & 2.22 & 3.06 & 4.64 \\ % \hline
  %   % \multicolumn{2}{c}{\textbf{FID with g.t.}} & 22.78 & 30.69 & 41.81 & 23.6 & 25.2 & 28.4 & xx & xx & xx \\ % \hline
  %   % \multicolumn{2}{c}{\textbf{FID with orig.}} & - & 8.41 & 19.81 & 1.67 & 3.11 & 6.09 & xx & xx & xx \\ % \hline
  %   \bottomrule
  %   \end{tabular}
  %   \parbox{0.9\textwidth}{\centering FID Scores for Parallel Methods on the Pixart and Flux.1. w/ G.T. means calculating the FID metrics with the ground-truth images. w/ Orig. means calculating the FID metrics with the images generated by the 1-Device original implementation.}
  %   % \caption{FID Scores for Different Models and Parallel Methods}
  %   % \label{tab:fid_scores}
  % \end{subfigure}
  \caption{Above: Showcases for 1024px Generation Images of PipeFusion and DistriFusion using Pixart. Bottom: FID Scores Evaluated for Pixart and Flux.1.
  We use a 20-step DPM-Solver with the warmup step set to 1 for DistriFusion and PipeFusion.
We use the COCO Captions 2014~\cite{chen2015microsoft} dataset to evaluate the FID scores. 
During the evaluation, a subset comprising 30,000 images is sampled from the validation set and resized to 256px to serve as the reference dataset. Concurrently, each experiment generates 30,000 images of 256px, each paired with a caption derived from the COCO Captions 2014 dataset, as the sample dataset. The quality of images generated by PipeFusion closely resembles that of the original images, regardless of whether 4 or 8 devices are used, and across varying patch numbers.
FID above the images is computed against the ground-truth images using Clean-FID~\cite{parmar2022aliased}.}
  \label{fig:quality}
\end{figure*}

In Figure~\ref{fig:quality}, we selected a few sample images generated using prompts from DistriFusion~\cite{li2023distrifusion} to help readers visually assess generation quality.
The images generated by PipeFusion are nearly indistinguishable from the original images to the human eye, across various configurations of the number of patches and devices.
Note that the FID evaluation results are independent of the specific prompts used for the visual demonstrations.

\section{Ablation Study}
\label{sec:ablstudy}
\begin{figure}[htbp!]
\centering
\includegraphics[width=1\textwidth]{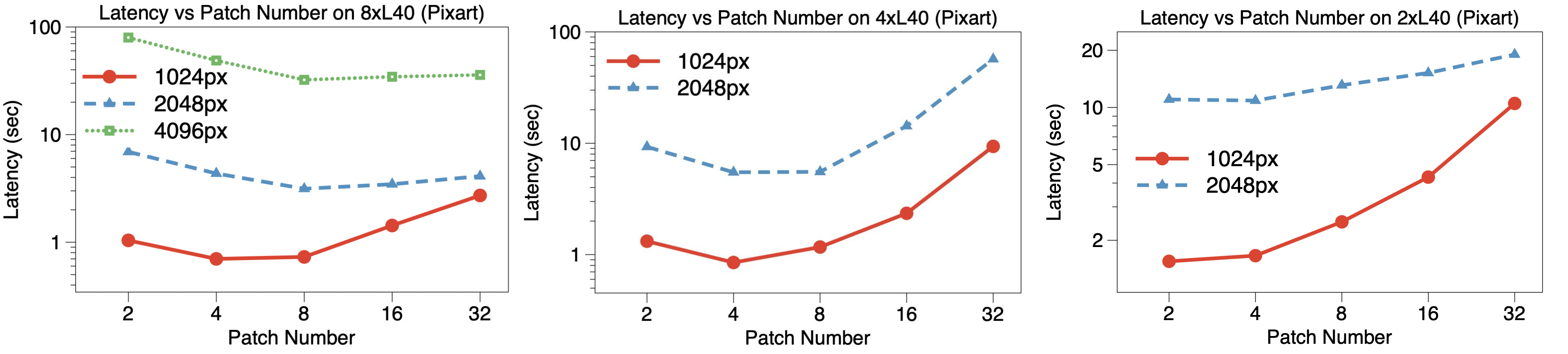}
\caption{Latency of PipeFusion (without warmup) for various patch numbers $M$ on 2x, 4x, and 8xL40 GPUs in image generation tasks at three different resolutions using Pixart.}
\label{fig:patchnumber}
\end{figure}

\textbf{Patch Number:} We analyze the impact of the setting of patch numbers $M$ on PipeFusion as shown in Figure~\ref{fig:patchnumber}. 
According to the analysis presented in Table~\ref{tab:parallel_schemes},  $M$ does not affect the communication cost.
However, as shown in the figure, the lowest latency is achieved when $M$ is equal to the number of GPUs, $N$.
If $M$ is large, the input size for operators becomes smaller, which can negatively impact the computational efficiency of the operator. 
In contrast, if $M$ is small, the efficiency of overlapping communication and computation decreases.

\textbf{Warmup Step}:
The number of warmup steps can negatively impact the overall performance of PipeFusion. 
The input temporal redundancy is relatively small during the initial few diffusion timesteps. 
Therefore, it is usually to employ a few warmup steps to enable synchronous communication without resorting to stale activations, which would introduce waiting time, commonly referred to as bubbles, into the pipeline.
As the number of sampling steps increases, the latency will also increase, as shown in Table~\ref{tab:warmp_impact}.
If the proportion of warmup steps is relatively small, no action is required. 
Our experiments involved a relatively small number of sampling steps, specifically 20 and 28. 
In other related work, it is common to use 50 steps~\cite{li2023distrifusion} or even 100 steps~\cite{ma2024deepcache}, which helps to alleviate the overhead of warmup.

\begin{table}[htbp!]
  \centering
  \begin{tabular}{cccccc}
  \toprule
  \multirow{2}{*}{Warmup} & \multicolumn{2}{c}{pixart} & \multicolumn{2}{c}{sd3-medium} & Flux.1-dev\\
  \cmidrule(lr){2-3}\cmidrule(lr){4-5}\cmidrule(lr){6-6}
  & pp=8 & cfg=2, pp=8 & pp=8 & cfg=2, pp=8 & pp=8 \\
  \midrule
  0 & 0.71 & 0.66 & 1.05 & 0.83 & 5.48\\
  \midrule
  1 & 0.82 & 0.69 & 1.16 & 0.87 & 5.53\\
    & (+15\%) & (+4\%) & (+10\%) & (+4\%) & (+1\%)\\
  \midrule
  2 & 0.91 & 0.70 & 1.27 & 0.92 & 6.00\\
    & (+28\%) & (+6\%) & (+21\%) & (+11\%) & (+9\%)\\
  \bottomrule
  \end{tabular}
  \caption{Impact of the warmup step number to Pixart, sd3-medium and Flux.1-dev on 1024px image generation, where pp indicates pipefusion (pp) parallel degree.}
  \label{tab:warmp_impact}
\end{table}

% \begin{table}[h!]
% \scriptsize
% \centering
% \begin{tabular}{cccccc}
% \toprule
% \multirow{2}{*}{Warmup} & \multicolumn{2}{c}{pixart} & \multicolumn{2}{c}{sd3-medium} & Flux.1-dev\\
% \cmidrule(lr){2-3}\cmidrule(lr){4-5}\cmidrule{6-6}
% & pp=8 & cfg=2, pp=8 & pp=8 & cfg=2, pp=8 & pp=8 \\
% \midrule
% 0 & 0.71 & 0.66 & 1.05 & 0.83 & 5.48\\

% 1 & 0.82 (+15\%) & 0.69 (+4\%) & 1.16 (+10\%) & 0.87 (+4\%) & 5.53 (+1\%) \\

% 2 & 0.91 (+28\%) & 0.7 (+6\%) & 1.27 (+21\%) & 0.92 (+11\%) & 6.00 (+9\%)\\
% \bottomrule
% \end{tabular}
% \caption{Impact of the warmup step number to Pixart, sd3-medium and Flux.1-dev on 1024px image generation, where pp indicates pipefusion (pp) parallel degree.}
% \label{tab:warmp_impact}
% \end{table}

To mitigate the performance degradation caused by warmup,
we can separate the warmup steps from the remaining working steps and allocate different computational resources to them. The output feature maps after the warmup steps can then be transmitted from the warmup devices to the working devices.
The warmup phase can apply sequence parallelism to fully utilize the computational resources.

We found that the setting of warmup steps is highly dependent on the model. 
By initializing the KV buffer to zero and not performing warmup, Pixart can still achieve satisfactory image generation results.
Through dynamic detection of the difference between the latent space input and the previous diffusion step, we can automatically set the warmup steps.
These optimizations for warm-up will be our future work.

\begin{table}[h]
\centering
\setlength{\tabcolsep}{5pt}
\begin{tabular}{lccc}
\toprule
Method & 1024px & 2048px & 4096px \\
\midrule
Tensor Parallel & 1.22 & 7.07 & 36.33 \\
DistriFusion & 0.77 & 2.48 & 25.24 \\
SP-Ring & 1.37 & 2.46 & 23.31 \\
SP-Ulysses & 2.59 & 3.54 & 27.41 \\
PipeFusion & \textbf{0.66} & \textbf{2.59} & \textbf{22.39} \\
\bottomrule
\end{tabular}
\caption{PixArt latency (sec) on 8$\times$A100 GPUs with NVLink interconnect. PipeFusion consistently delivers the lowest latency.}
\label{tab:a100-nvlink}
\end{table}

\textbf{Evaluation on NVLink connected GPUs:}
To evaluate the sensitivity of PipeFusion to hardware interconnect bandwidth, we conducted additional experiments on 8$\times$A100 GPUs connected via NVLink (high-bandwidth) in contrast to the default 8$\times$L40 PCIe setting. Table~\ref{tab:a100-nvlink} reports the latency for PixArt inference across different resolutions. PipeFusion consistently achieves the lowest latency across all resolutions and remains robust even under high-bandwidth NVLink settings. While the relative performance gap narrows at higher resolutions due to the increasing dominance of computation cost (i.e., the shift from communication- to computation-bound regimes), PipeFusion is still competitive or superior to all baselines. We also emphasize that the FID score is invariant to GPU type or interconnect bandwidth, as it is solely determined by the model architecture and inference algorithm.

%%%%%%%%%%%%%%%%%%%%%%%%%%%%%%%%%%%%%%%%%%%%%%%%%%%%%%%%%%%%

\newpage

\end{document}